\pdfoutput=1

\documentclass[11pt]{article}

\usepackage[final]{acl}

\usepackage{times}
\usepackage{latexsym}
\usepackage{amsmath}

\usepackage[T1]{fontenc}

\usepackage[utf8]{inputenc}

\usepackage{microtype}

\usepackage{inconsolata}

\usepackage{graphicx}
\usepackage{booktabs}
\usepackage{arydshln}
\usepackage{makecell}
\usepackage{pifont}
\usepackage{fontawesome5}
\usepackage{svg}
\newcommand{\myxmark}{\ding{55}}
\newcommand{\mychecking}{\ding{51}}

\title{MAVL: A Multilingual Audio-Video Lyrics Dataset for Animated Song Translation }

\author{
 \textbf{Woohyun Cho\textsuperscript{1}},
 \textbf{Youngmin Kim\textsuperscript{1}},
 \textbf{Sunghyun Lee\textsuperscript{1}},
 \textbf{Youngjae Yu\textsuperscript{2}\thanks{Corresponding Author}},
\\
 \textsuperscript{1}Yonsei University \quad 
 \textsuperscript{2}Seoul National University
\\
   \texttt{\href{mailto:k106419@yonsei.ac.kr}{k106419@yonsei.ac.kr}} \quad
   \texttt{\href{mailto:youngjaeyu@snu.ac.kr}{youngjaeyu@snu.ac.kr}}
}

\usepackage[utf8]{inputenc} 
\usepackage[T1]{fontenc}    
\usepackage{kotex}          
\usepackage{xcolor}

\newcommand{\stablets}{\footnote{https://github.com/jianfch/stable-ts}}

\newcommand{\lyricstranslate}{\footnote{https://lyricstranslate.com/}}

\usepackage{array}
\usepackage{multirow}
\usepackage{CJKutf8}

\begin{document}
\maketitle
\begin{abstract}
Lyrics translation requires both accurate semantic transfer and preservation of musical rhythm, syllabic structure, and poetic style. In animated musicals, the challenge intensifies due to alignment with visual and auditory cues. We introduce Multilingual Audio-Video Lyrics Benchmark for Animated Song Translation (MAVL), the first multilingual, multimodal benchmark for singable lyrics translation. By integrating text, audio, and video, MAVL enables richer and more expressive translations than text-only approaches. Building on this, we propose Syllable-Constrained Audio-Video LLM with Chain-of-Thought (\textbf{\texttt{SylAVL-CoT}}), which leverages audio-video cues and enforces syllabic constraints to produce natural-sounding lyrics. Experimental results demonstrate that \textbf{\texttt{SylAVL-CoT}} significantly outperforms text-based models in singability and contextual accuracy, emphasizing the value of multimodal, multilingual approaches for lyrics translation. 
\newcounter{savefootnote}
\setcounter{savefootnote}{\value{footnote}}  
\begingroup
\makeatletter
\renewcommand\@makefntext[1]{%
    \noindent\@makefnmark\hspace{0.5em}#1}
\makeatother
\renewcommand{\thefootnote}{}
\footnote{\faGithub\hspace{0.1em}Code: \href{https://github.com/k1064190/MAVL}{k1064190/MAVL}}
\footnote{\includegraphics[width=1.0em]{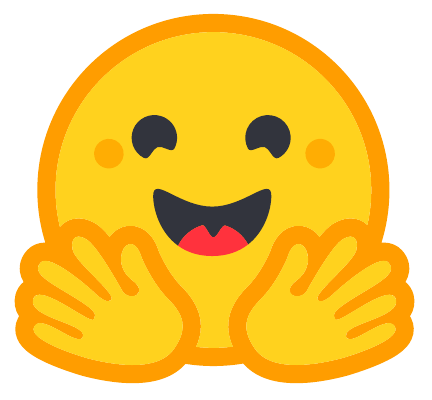}\hspace{0.1em}Dataset: \href{https://huggingface.co/datasets/Noename/MAVL}{NoeName/MAVL}}
\endgroup
\setcounter{footnote}{\value{savefootnote}}  
\end{abstract}

\section{Introduction}
\begin{figure*}[ht!]
    \centering
    \includegraphics[width=\linewidth]{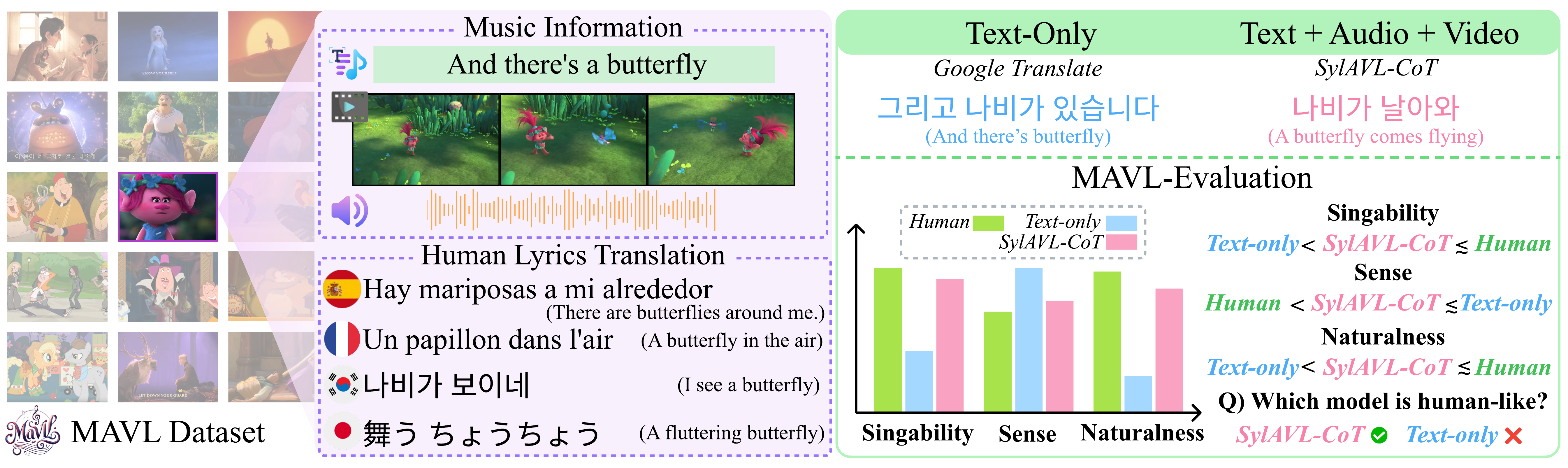}
    \caption{\textbf{Overview of MAVL Benchmark.} This lyric example is part of the OST ``Get Back up Again'' from ``Troll'', produced by Disney Corporation. the left illustrates MAVL Dataset components (music, multilingual human lyrics). The right compares translation outputs, showing our audio-visual \textbf{\texttt{SylAVL-CoT}} produces more vivid and human-like results than text-only models.}
    \label{fig:teaser_figure}
\end{figure*}
Lyric translation, a specialized task, prioritizes “singability”—fitting lyrics to melody. This often requires beyond-literal translation to preserve both musicality and meaning, making it significantly more complex than standard text translation.

However, text-based lyric translation has inherent limitations, especially in musical animations. For example, consider the lyric ``And there's a butterfly'' from the song ``Get Back Up Again'' in the movie Trolls. A text-only translation, such as one provided by Google Translate, might yield ``그리고 나비가 있습니다'' (Geu-ri-go na-bi-ga it-seum-ni-da), which literally means ``And there's a butterfly.'' While this conveys the basic presence of a butterfly, it lacks the dynamic action depicted visually and offers poor singability. In contrast, by incorporating audio and video context, a multimodal system like \textbf{\texttt{SylAVL-CoT}} can produce a translation such as ``나비가 날아와'' (Na-bi-ga na-ra-wa), meaning ``A butterfly comes flying''. This version, as shown in Figure~\ref{fig:teaser_figure}, is more vivid, aligns with the on-screen motion, and demonstrates superior singability, naturalness, and human-likeness compared to the text-only approach. Appendix~\ref{appendix:reasoning_trajectory} details \textbf{\texttt{SylAVL-CoT}}'s multimodal reasoning for context-aware translations, essential for musical and cinematic cohesion through rhythm and visual storytelling. Furthermore, cross-lingual syllabic and rhythmic differences necessitate adaptations beyond literal translation, incorporating musical elements for naturalness and markedly increasing complexity over standard text translation.

However, despite this importance, previous studies rely on text-based or text-and-score-based approaches, limited by musical constraints and specific languages~\cite{guo2022automaticsongtranslationtonal, kim2024kpoplyrictranslationdataset, li2023translatebeautysongsjointly, ye2024singitnarrateit}.
To address these limitations, we introduce the MAVL, \textbf{M}ultilingual \textbf{A}udio-\textbf{V}ideo \textbf{L}yrics Benchmark for Animated Song Translation. MAVL is a novel benchmark for multilingual, multimodal lyric translation in animated musicals (Figure~\ref{fig:teaser_figure}), featuring aligned lyrics (English, Spanish, French, Korean, and Japanese) with audio-video data. This allows models to integrate textual, auditory, and visual information for more contextually and emotionally resonant translations.

Leveraging MAVL effectively requires models that jointly process text, audio, and video while maintaining linguistic and musical coherence. To address current limitations in handling such multimodal data, we propose \textbf{Syl}lable constrained \textbf{A}udio-\textbf{V}ideo \textbf{L}LM with \textbf{C}hain \textbf{o}f \textbf{T}hought (\textbf{\texttt{SylAVL-CoT}}). \textbf{\texttt{SylAVL-CoT}} enhances standard reasoning by incorporating audio and video cues, enabling better integration of contextual information across modalities for more accurate and musically coherent lyric translations.

Furthermore, to systematically evaluate multilingual, multimodal lyric translation models leveraging MAVL, we propose evaluation metrics for more human-like results. Using these metrics, we analyze MAVL, quantitatively and qualitatively evaluate our approach against existing models, and conduct ablation studies demonstrating the necessity of multimodal input and chain-of-thought stages for effective lyric translation.

In summary, our main contributions are:
\begin{itemize}
\item We introduce \textbf{MAVL}, a multilingual, multimodal dataset and benchmark for multimodal lyric translation in animated musicals, consisting of aligned text, audio, and video data across five languages.
\item We establish evaluation metrics for multilingual lyrics translation.
\item We propose a \textbf{\texttt{SylAVL-CoT}} for lyric translation that enhances standard reasoning by incorporating audio and video cues.
\end{itemize}

\begin{figure*}[thb!]
    \centering
    \includegraphics[width=\linewidth]{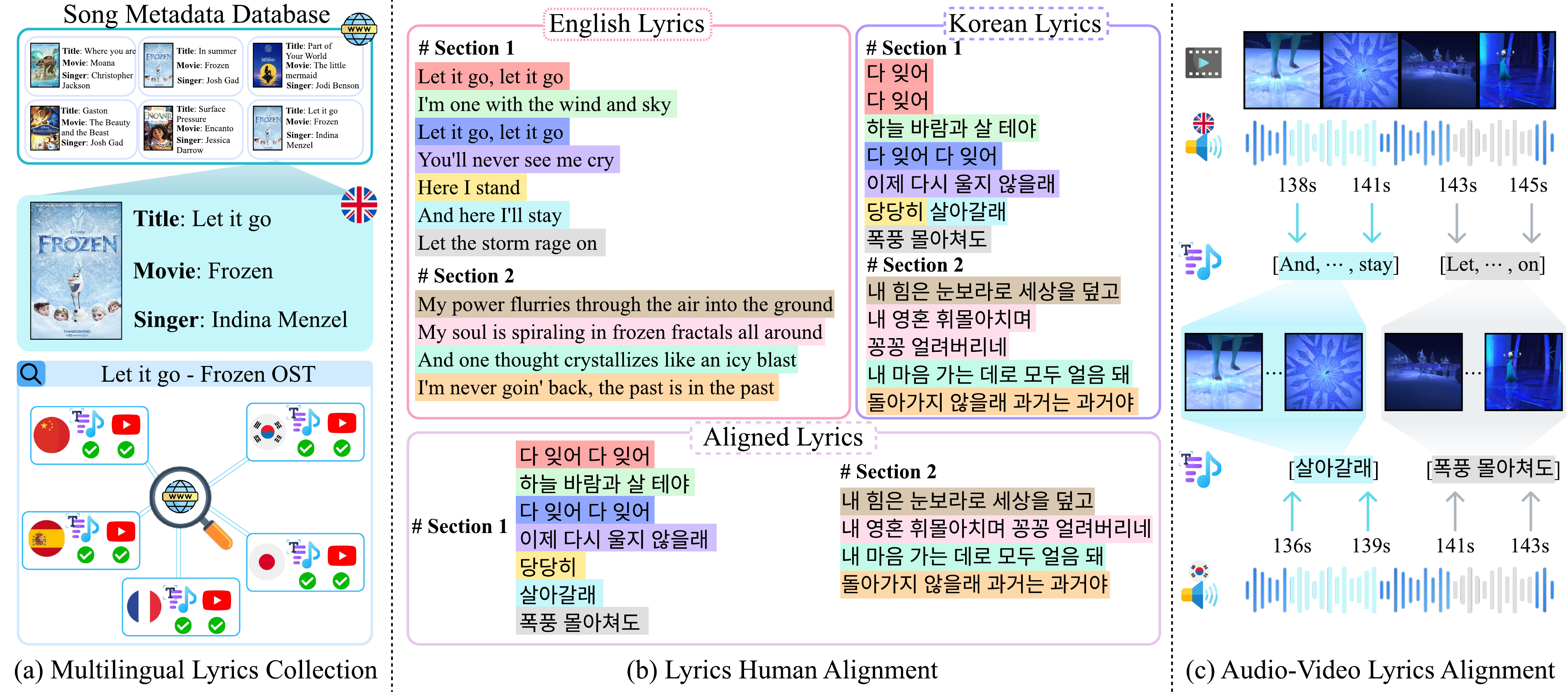}
    \caption{\textbf{MAVL dataset collection pipeline.} (b) visualizes the lyric alignment process, where each color corresponds to English and Korean, respectively. This example lyrics and images are part of the OST ``Let it go'' from Frozen, produced by Disney Corporation. For more details, refer to Section~\ref{sec:dataset}}
    \label{fig:dataset_collection}
\end{figure*}

\section{Related works}
\noindent \textbf{Lyrics Translation Challenges and Strategies in Translation Studies. }
Translating lyrics has long been recognized as a specialized domain in translation studies, as it must balance semantic equivalence, poetic structure, and musical requirements~\cite{franzon2008choices, low2003singable, low2005pentathlon}. 
Early frameworks propose strategies ranging from literal translation to complete adaptation, guided by the ``Pentathlon Principle'' (singability, sense, naturalness, rhythm, and rhyme). 
Research on musicals and Disney soundtracks highlights deliberate manipulation of rhyme schemes and syllable counts to maintain musical flow, alongside cultural shifts for humor and emotional nuance \cite{Leni2019ANALYZINGTS, susam2008translation}. 
In audio-visual translation (AVT), filmic elements such as camera angles, music tempo, and background music, especially in musicals and animations is important \cite{banos2009prefabricated, 1075843ar, taylor2016multimodal, Pidhrushna2021FunctionalAT}. Certain scenes rely heavily on imagery or character expressions to convey emotional subtext \cite{supardi2018audio}, making strict fidelity to source lyrics potentially mismatched with the visual narrative.

\noindent \textbf{Lyrics Machine Translation. }
Building on the principle that lyric translation prioritizes “singability,” recent work has advanced the field through dataset creation, joint learning of melodic and textual features, and development of specialized evaluation metrics.
For instance, ~\cite{guo2022automaticsongtranslationtonal} incorporated tonal constraints for translating lyrics into Mandarin, while \cite{ou2023songsborderssingablecontrollable} experimented with integrating melody-length and phonetic constraints into translation systems. Also, \cite{kim2023computationalevaluationframeworksingable,kim2024kpoplyrictranslationdataset} developed a K-pop lyric translation dataset and trained a model. Further advancements include \cite{li2023translatebeautysongsjointly}, who presented a method for jointly learning melody and lyric semantics, and \cite{ye2024singitnarrateit}, which demonstrated a system that successfully fuses semantic fidelity with musical coherence.

\noindent \textbf{Multimodal Chain-of-Thought Reasoning. }
Multimodal Chain-of-Thought (CoT) reasoning~\cite{zhang2024multimodalchainofthoughtreasoninglanguage, ma2025audiocotexploringchainofthoughtreasoning} extends traditional CoT prompting~\cite{wei2022:chain} by incorporating vision modality to enhance complex reasoning tasks~\cite{wang2024:videocot,Xu2024LLaVACoTLV}.
Moreover, CoT prompting has been utilized to generalize reasoning across diverse languages and specialized knowledge domains~\cite{hu2024largelanguagemodelmultidomain}. 

Beyond these advancements, CoT prompting has been utilized in Multimodal machine translation tasks~\cite{barrault2018:findings} to improve translation quality~\cite{rajpoot2024:multimodal, he2024:multi} and has also been applied to machine translation evaluation~\cite{qian2024:large}. Building on this, multimodal CoT techniques have also been explored in speech translation~\cite{du2024cotstenhancingllmbasedspeech, hu2024:chain}.

\section{Multilingual Audio-Video for Lyrics Translation Benchmark}
We introduce MAVL, the Multilingual Audio-Video Lyrics Benchmark for Animated Song Translation, designed to evaluate lyric translation models integrating text, audio, and video, ensuring linguistic and musical coherence.
MAVL comprises three key components: 1) a dataset of aligned lyrics, audio, and video (Section~\ref{sec:dataset}), 2) a syllable-constrained audio-video LLM with the chain-of-thought called \textbf{\texttt{SylAVL-CoT}} (Section~\ref{sec:multimodal_cot}), and 3) evaluation metrics (Section~\ref{sec:metrics}). 
\begin{table}[tbp]
\centering
\resizebox{\columnwidth}{!}{%
\begin{tabular}{cccccc}
\toprule
Datasets & Lang & Songs & Modality & Singablilty & Available \\ \midrule
\multicolumn{1}{c|}{\cite{li2023translatebeautysongsjointly}}   & En, Zh & 79  & Text, Score & \mychecking & \myxmark \\
\multicolumn{1}{c|}{\cite{kim2024kpoplyrictranslationdataset}}     &  En, Ko  & 1,000 & Text & \mychecking & \mychecking \\
\multicolumn{1}{c|}{\cite{ye2024singitnarrateit}}     &  En, Zh  & 11$^{*}$ & Text & \myxmark & \myxmark \\
\multicolumn{1}{c|}{\cite{ou2023songsborderssingablecontrollable}}                                                      & En, Zh & 5,341  & Text & \myxmark & \mychecking    \\
\multicolumn{1}{c|}{MAVL}  & En, Fr, Es, Ko, Ja  & 228 & Text, Video, Audio  & \mychecking & \mychecking  \\ \bottomrule
\end{tabular}%
}
\caption{\textbf{Comparison of lyrics translation parallel datasets.} ``*'' indicates the number of musicals. The number of songs per musical is not specified in the respective paper.}
\label{table:comp_stats}
\end{table}

\subsection{Multilingual Audio-Video for Lyrics Translation Dataset}
\label{sec:dataset}
This section details our MAVL dataset collection pipeline, illustrated in Figure~\ref{fig:dataset_collection}. The dataset contains lyrics and corresponding audio-video data for 228 songs across five languages (English, Spanish, French, Japanese, and Korean), as summarized in Table~\ref{table:lyrics_stats}. To our knowledge, MAVL is the first dataset to support multilingual lyric translation across three modalities. Further details are in Appendix~\ref{appendix:appendix_dataset}.

\noindent \textbf{Multilingual Lyrics Collection. }
We first gathered metadata (song titles, artists) for English animated film music from \texttt{last.fm}. Original English lyrics were then retrieved from \texttt{genius} and manually verified against the songs.
Based on these verified English song titles, we proceeded to collect non-English lyrics corresponding to official dubbed versions of the original English songs in four languages: Spanish, French, Korean, and Japanese. We utilized platforms such as lyricstranslate\lyricstranslate{}. Searching this platform by the original English song title typically yields a list of that song's lyrics in various languages. These listed versions often represent the official, singable dubbed interpretations, distinct from literal, non-singable translations that might be found if browsing general ``translation'' categories. Alongside these candidate lyrics, we gathered the corresponding audio and video.

\noindent \textbf{Lyrics Human Alignment. }
The human alignment process was crucial not only for segmenting lyrics but also for rigorously verifying their authenticity as official, singable dubbed versions. This involved simultaneously cross-referencing the original English audio/video with the non-English candidate lyrics and their corresponding official dubbed audio/video.  During this stage, candidate non-English lyrics were critically evaluated: if a set of lyrics could not be confirmed against an official audio-visual release, or if they did not accurately match the sung content in the verified official dub, that specific language version was excluded from our dataset. This process ensured that only verified, officially dubbed lyrics were retained.
During alignment, non-singable dialogue or overlapping lyrics (prioritizing main melody) were also excluded.


\begin{table}[tbp]
\centering
\setlength{\tabcolsep}{4pt} 
\resizebox{\columnwidth}{!}{%
\begin{tabular}{ccccc}
\toprule
Language                                                                          & \# Songs & \# Video & \# Sections & \# Lines \\ \midrule
\multicolumn{1}{c|}{English}   & $228$      & $228$   & $1,923$        & $6,623$     \\[0.5ex]
\cdashline{1-5}
\noalign{\vskip 0.5ex}
\multicolumn{1}{c|}{Spanish}                                                      & $201$      &   $181$    & $1,595$        & $5,739$     \\
\multicolumn{1}{c|}{French}                                                       & $158$      &   $143$    & $1,421$        & $4,821$     \\
\multicolumn{1}{c|}{Japanese}                                                     & $138$      &   $114$    & $1,264$        & $4,280$     \\
\multicolumn{1}{c|}{Korean}                                                       & $133$      &   $117$    & $1,138$        & $3,974$     \\ \bottomrule
\end{tabular}%
}
\caption{\textbf{Statistics of the MAVL Benchmark Dataset.} ``\# Section'' refers to sections of the lyrics, while ``Lines'' denotes the individual lines within those sections. The number of videos is equal to the number of audio.}
\label{table:lyrics_stats}
\end{table}

\noindent \textbf{Audio-Video-Lyrics Alignment. }
To align audio, video, and lyrics, we utilized \texttt{stable-ts}\stablets{}, a Whisper model~\cite{radford2022robustspeechrecognitionlargescale}-based tool, to generate stable timestamps. This allowed us to determine the start and end times of each lyric line and segment the corresponding audio and video, extracting synchronized audio and video information aligned with the lyrics. Detailed alignment methodology, including our ensemble approach and quality assurance process, is provided in Appendix~\ref{appendix:alignment_methodology}.

\begin{figure}
    \centering
    \includegraphics[width=\linewidth]{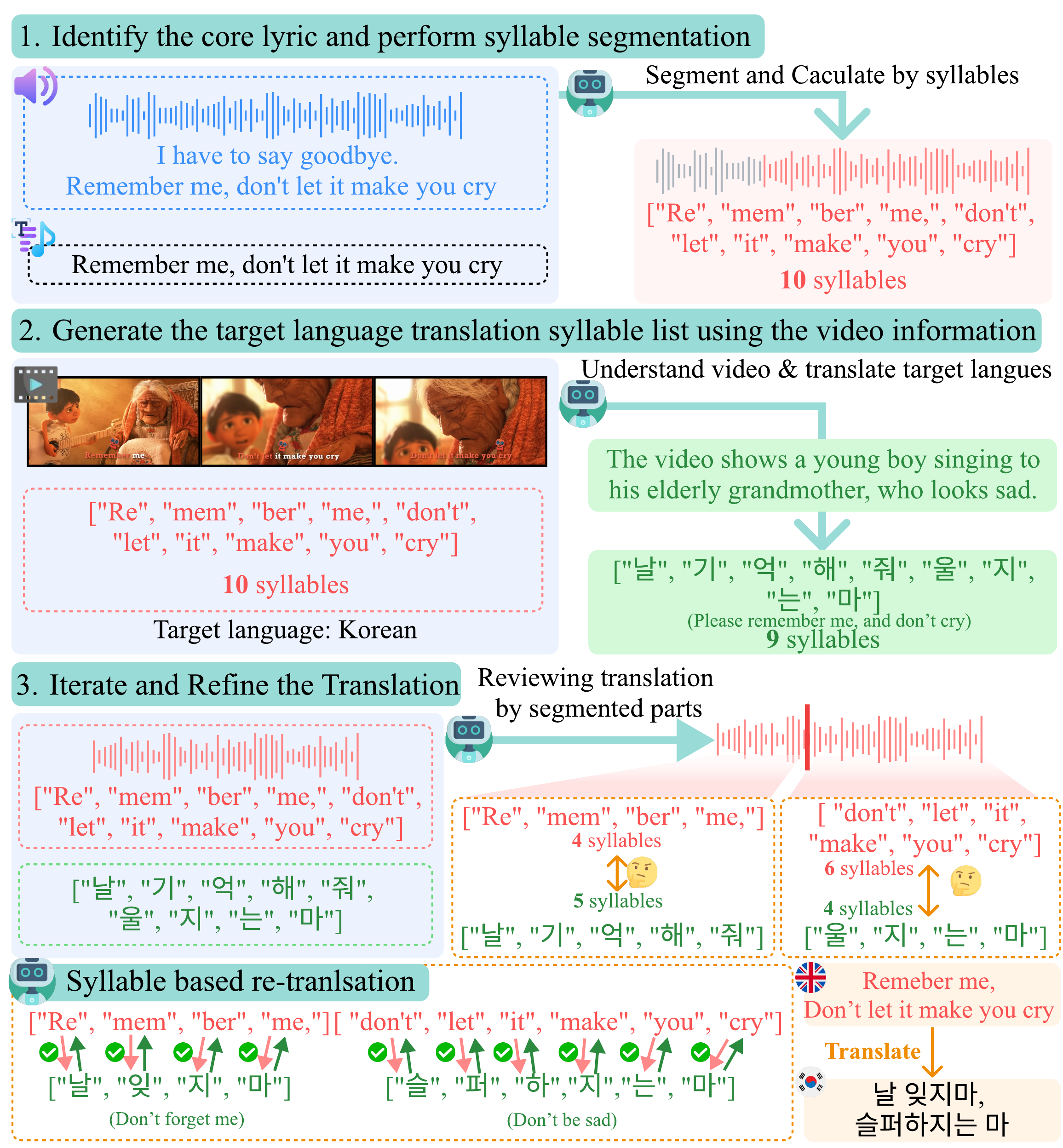}
    \caption{\textbf{\textbf{\texttt{SylAVL-CoT}} pipeline for lyrics translation.} This three-step process segments syllables utilizing audio, translates using video context, and iteratively refines the output to match original syllable counts.}
    \label{fig:sylavl-cot_pipeline}
\end{figure}

\subsection{Syllable-Constrained Audio-Video LLM with the Chain of Thought}
\label{sec:multimodal_cot}
In this section, we introduce Syllable-Constrained Audio-Video LLM with Chain of Thought (\textbf{\texttt{SylAVL-CoT}}), a multimodal approach designed to enhance lyrics translation by integrating audio, video, and text while maintaining rhythmic and semantic coherence.
Existing Multilingual Large Language Models (MLLMs) struggle to integrate audio, video, and text, and the lack of aligned multilingual audio-visual datasets makes fine-tuning impractical. Additionally, they fail to maintain syllabic alignment, leading to unnatural phrasing and disrupted singability in translated lyrics.

To address this, we leverage Gemini 2.0 Flash~\cite{gemini20flash} and adopt Chain-of-Thought (CoT) reasoning~\cite{wei2022:chain} to incorporate multimodal cues and enforce syllable constraints.
Specifically, we implement syllable-aware alignment, where the model dynamically adjusts phrasing to match the original syllable count and rhythmic structure. The syllable constraint is applied during inference by providing the model with the exact syllable count of the source lyrics via prompting, instructing it to generate translations that match this count as closely as possible (see Appendix~\ref{appendix:cot_prompts} for detailed prompts). 
Our \textbf{\texttt{SylAVL-CoT}} follows three-step process, which is illustrated in Figure~\ref{fig:sylavl-cot_pipeline}. Examples of the model's detailed reasoning trajectories for lyrics translation are provided in Appendix~\ref{appendix:reasoning_trajectory}.

\noindent \textbf{Identify the Core Lyric and Perform Syllable Segmentation. } We begin by supplying the model with a specific segment of the original lyric text, accompanied by an audio snippet. The model’s task is to locate precisely which part of the audio is relevant to the text. Next, the system carefully segments the lyric into syllables based on audible breaks in the audio—an essential first step for maintaining the original rhythm and singability.
This initial segmentation guides subsequent steps, providing a structural template for the translation.

\noindent \textbf{Generate the Target-Language Translation Syllable List, Utilizing Video Context. } In the second stage, the pipeline processes visual cues (e.g., thematic elements, animation style, cultural context) from the video to refine the translation. The model aims to capture not only the literal meaning of the lyrics but also subtleties related to imagery, cultural nuances, and artistic style.
During this step, the model strives to preserve the original syllable count to maintain or approximate the musical flow.

\noindent \textbf{Iterate and Refine the Translation. } Preserving syllable count, natural flow, and rhythmic fidelity can be challenging—especially when translating between languages with differing grammatical structures and phonetic inventories. To handle this complexity, the model iteratively refines the translated text by paraphrasing or reordering words until it achieves a final output that is both culturally appropriate and linguistically coherent. The model checks the number of syllables against the target and, if needed, continues to adjust the translation or segmentation.

\begin{table*}[!t]
\centering
\scriptsize
\setlength{\tabcolsep}{4pt}
\renewcommand{\arraystretch}{1.2}
\resizebox{\textwidth}{!}{%
\begin{tabular}{p{0.5cm} c *{12}{>{\centering\arraybackslash}p{0.8cm}} }
\toprule
\multirow{2}{*}{} & \multirow{2}{*}{Model} & \multicolumn{4}{c}{Syllable Error (SE) $\downarrow$} & \multicolumn{4}{c}{Syllable Distance$\downarrow$} & \multicolumn{4}{c}{Error Rate$\downarrow$} \\
\cmidrule(lr){3-6}\cmidrule(lr){7-10}\cmidrule(lr){11-14}
                  &                        & ES   & FR   & KO   & JA   & ES   & FR   & KO   & JA   & ES   & FR   & KO   & JA   \\
\midrule
\multicolumn{2}{c}{} & \multicolumn{12}{c}{{\textbf{English Lyrics $\longleftrightarrow$ Translated Lyrics}}} \\
\midrule
\multirow{1}{*}{} & Human Expert          & 1.780 & 1.031 & 1.052 & 1.716 & 0.115 & 0.098 & 0.141 & 0.096 & 0.587 & 0.455 & 0.353 & 0.607 \\[0.5ex]
\cdashline{1-14}
\noalign{\vskip 0.5ex}
\multirow{5}{*}{\rotatebox{90}{w/o SC}} 
                 & Google Translate      & 7.784 & 4.140 & 12.226 & 20.924 & 0.411 & 0.248 & 0.619 & 0.951 & 0.926 & 0.818 & 0.943 & 0.975 \\
                 & mBART-50              & 25.790 & 4.397 & 13.036 & 15.617 & 1.443 & 0.268 & 0.650 & 0.741 & 0.953 & 0.831 & 0.954 & 0.956 \\
                 & Qwen2.5-72B              & 8.063 & 4.132 & 9.870 & 15.654  & 0.426 & 0.249 & 0.515 & 0.750 & 0.939 & 0.816 & 0.939 & 0.973 \\
                 & GPT-4o                & 8.100 & 4.158 & 10.302 & 16.711  & 0.420 & 0.245 & 0.524 & 0.780 & 0.950 & 0.824 & 0.947 & 0.984 \\
                 & Gemini 2.0                & 8.718 & 4.660 & 10.819 & 16.919  & 0.456 & 0.276 & 0.557 & 0.798 & 0.948 & 0.829 & 0.944 & 0.977 \\
\cdashline{1-14}
\multirow{4}{*}{\rotatebox{90}{w/ SC}} 
                 & Qwen2.5-72B              &   6.512   &   2.729   &   6.502   &   10.061   & 0.354 & 0.183 & 0.361 & 0.517 & 0.927 & 0.754 & 0.911 & 0.962 \\
                 & GPT-4o                &   3.164   &   1.442   &   3.084    &   7.221   & 0.185 & 0.138 & 0.182 & 0.378 & 0.795 & 0.676 & 0.935 & 0.935 \\
                 & Gemini 2.0                & 3.585 & 1.384 & 3.039 & 6.257  & 0.207 & 0.104 & 0.190 & 0.389 & 0.872 & 0.604 & 0.848 & 0.952 \\
                 & SylAVL-CoT (Gemini 2.0)         & \underline{0.966} & \textbf{0.902} & \underline{0.695} & \underline{2.572} & \underline{0.063} & \underline{0.089} & \underline{0.049} & \underline{0.155} & \underline{0.352} & \underline{0.510} & \underline{0.237} & \underline{0.611} \\
                 & SylAVL-CoT (Gemini 2.5)   & \textbf{0.918} & \underline{0.909} & \textbf{0.660} & \textbf{2.507} & \textbf{0.059} & \textbf{0.086} & \textbf{0.046} & \textbf{0.151} & \textbf{0.334} & \textbf{0.498} & \textbf{0.225} & \textbf{0.596} \\
\midrule
\multicolumn{2}{c}{} & \multicolumn{12}{c}{\textbf{Dubbed Lyrics $\longleftrightarrow$ Translated Lyrics}} \\
\midrule
\multirow{5}{*}{\rotatebox{90}{w/o SC}} 
                 & Google Translate      & 6.513 & 4.157 & 11.848 & 20.398 & 0.340 & 0.272 & 0.624 & 0.928 & 0.886 & 0.806 & 0.944 & 0.967 \\
                 & mBART-50              & 24.481 & 4.424 & 12.651 & 15.182  & 0.126 & 0.292 & 0.655 & 0.729 & 0.929 & 0.837 & 0.950 & 0.945 \\
                 & Qwen2.5-72B              & 6.745 & 4.158 & 9.518 & 15.165  & 0.351 & 0.275 & 0.523 & 0.928 & 0.900 & 0.807 & 0.936 & 0.964 \\
                 & GPT-4o                & 6.717 & 4.237 & 9.907 & 16.227  & 0.341 & 0.289 & 0.538 & 0.768 & 0.894 & 0.814 & 0.931 & 0.974 \\
                 & Gemini 2.0                & 7.347 & 4.658 & 10.436 & 16.364  & 0.377 & 0.299 & 0.560 & 0.772 & 0.908 & 0.831 & 0.936 & 0.964 \\
\cdashline{1-14}
\multirow{4}{*}{\rotatebox{90}{w/ SC}} 
                 & Qwen2.5-72B              &   5.185  &   2.893   &   6.229   &   9.703   & 0.279 & 0.222 & 0.370 & 0.509 & 0.876 & 0.758 & 0.898 & 0.942 \\
                 & GPT-4o                &   2.403   &   1.723   &   3.067   &   6.940   & 0.162 & 0.183 & 0.212 & 0.338 & 0.742 & 0.709 & 0.906 & 0.906 \\
                 & Gemini 2.0                & 2.610 & 1.703 & 2.941 & 5.996  & 0.163 & 0.150 & 0.205 & 0.382 & 0.759 & 0.667 & 0.813 & 0.917 \\
                 & SylAVL-CoT (Gemini 2.0)         & \underline{1.349} & \underline{1.363} & \underline{1.261} & \underline{3.107} & \underline{0.125} & \underline{0.149} & \underline{0.122} & \underline{0.223} & \underline{0.634} & \underline{0.631} & \underline{0.472} & \underline{0.759} \\
                 & SylAVL-CoT (Gemini 2.5)   & \textbf{1.299} & \textbf{1.269} & \textbf{1.200} & \textbf{2.979} & \textbf{0.120} & \textbf{0.139} & \textbf{0.116} & \textbf{0.214} & \textbf{0.611} & \textbf{0.603} & \textbf{0.449} & \textbf{0.728} \\
\bottomrule
\end{tabular}%
}
\caption{\textbf{Comparison with other models: Syllable errors.} ``SC'' means syllable constraint. The syllable constraint is a condition applied when using LLM models. If the syllable constraint is not applied, the model is prompted to perform a simple translation. Conversely, when the syllable constraint is applied, the syllable count from the original lyrics is provided as a condition. \textbf{Best} and \underline{second} are highlighted.}
\label{tab:comp_syllable}
\end{table*}

\subsection{Evaluation Metrics}
\label{sec:metrics}
We evaluate lyric translation quality across three principles.
\textit{Singability} evaluates whether the translated lyrics are suitable for singing, \textit{Sense} is about accurately conveying the meaning and message of the original lyrics, and \textit{Naturalness} evaluates whether the translated lyrics sound natural and conversational in the target language. Details are in Appendix~\ref{appendix:metrics_selection}.

\noindent \textbf{Syllable Error. }
Existing lyric translation evaluation metrics~\cite{kim2023computationalevaluationframeworksingable} primarily rely on the syllable count of the original language, failing to capture the subtle nuances of multilingual lyric translation, which can negatively impact \textit{Singability}.
In particular, simply comparing syllable counts without considering phonological and rhythmic differences across languages risks compromising the naturalness of translated lyrics. Therefore, it is essential to incorporate dubbed lyrics that adapt syllable counts.

Syllable error (SE) measures how well the syllable count aligns with the original English lyrics ($c_{en}$) and how closely it matches the dubbed lyrics, which reflect the linguistic characteristics of the target language ($c_{dub}$).
For the purpose of explanation, we will denote both $c_{en}$ and $c_{dub}$ as $c$. 
Given the syllable counts, $c$ and syllable counts of machine-translated text, $c_{pred}$, syllable error (SE) is calculated as follows.
\begin{equation}
    SE =
    \begin{cases}
        c - c_{pred}, & \text{if } c \geq c_{pred} \\
        \beta (c_{pred} - c), & \text{if } c < c_{pred}
    \end{cases}
\end{equation}
where $\beta \ge 1$ is a penalty factor for exceeding the reference syllable count. We set the additional penalty $\beta = 2.0$ in our experiments as suggested by~\cite{ye2024singitnarrateit}, to penalize exceeding the syllable count more heavily, as it can be more detrimental to \textit{Singability}.
In addition, we also employ the Syllable Count Distance (SCD) error rate proposed in \cite{kim2024kpoplyrictranslationdataset}.
SCD is defined as:
\begin{equation}
    SCD = \frac{1}{2} \left(\frac{\lvert c - c_{pred}\rvert}{c} + \frac{\lvert c - c_{pred}\rvert}{c_{pred}}\right) 
\end{equation}
Finally, the error rate is defined as the proportion of lines whose predicted syllable counts do not match the reference (original or dubbed) syllable counts.

\noindent \textbf{Semantic scores.}
Accurately conveying the semantic meaning of lyrics is crucial in lyric translation. While previous studies \cite{li2023translatebeautysongsjointly, ou2023songsborderssingablecontrollable} primarily relied on word-overlap-based evaluation metrics such as BLEU \cite{papineni2002:bleu} and TER \cite{snover-etal-2006-study} to assess semantic similarity, these approaches have clear limitations when translating creative content that demands both fluency and inventiveness.

To address these limits, we utilize cosine similarity measures using multilingual embeddings from pre-trained sentence embedding models, namely mpnet-base-v2~\cite{song2020mpnetmaskedpermutedpretraining}, as follows:
\begin{equation}
    sim_{M} = \text{cos}(e^{M}_{gt}, e^{M}_{pred}),
\end{equation}
where, $M$ represents the model (mpnet-base-v2), $e_{gt}$ and $e_{pred}$ denotes the model's embeddings of ground truth and machine-translated lyrics, and $\text{cos}(\cdot, \cdot)$ refers to the cosine similarity function. We treat both the original English lyrics and the dubbed lyrics as ground truths. 
We also performed experiments using other multilingual embedding models in Appendix~\ref{appendix:additional_exp}.

\noindent \textbf{Phonetic Distance.}
We evaluate \textit{Naturalness} by assessing pronunciation similarity.
To do this, we convert the lyrics into International Phonetic Alphabet (IPA)~\cite{laver1994:principles} and measure the phonetic similarity using Levenshtein distance~\cite{levenshtein1966:binary}, as follows:
\begin{equation}
    D = \text{LD}(\text{IPA}_{gt}, \text{IPA}_{pred}),
\end{equation}
where $\text{LD}(\cdot, \cdot)$ refers to the Levenshtein distance function.

\begin{table}[t]
    \centering
    \scriptsize
    \setlength{\tabcolsep}{8pt} 
    \resizebox{\columnwidth}{!}{%
    \begin{tabular}{p{0.5cm} l *{4}{>{\centering\arraybackslash}p{0.7cm}}} 
    \toprule
    \multirow{2}{*}{} & \multirow{2}{*}{Model} & \multicolumn{4}{c}{MPNet$\uparrow$} \\
    \cmidrule(lr){3-6} 
    &       & ES   & FR   & KO   & JA \\
    \midrule
    & & \multicolumn{4}{c}{English Lyrics $\longleftrightarrow$ Translated} \\
    \midrule
    \multirow{1}{*}{} & Human Expert   & 0.639 & 0.613 & 0.575 & 0.537 \\[0.5ex]
    \cdashline{1-6} 
    \noalign{\vskip 0.5ex}
    \multirow{5}{*}{\rotatebox{90}{w/o SC}}
    & Google Translate & \textbf{0.905} & \textbf{0.903} & \textbf{0.857} & \textbf{0.857} \\
    & mBART-50       & 0.766 & 0.890 & 0.850 & 0.842 \\
    & Qwen2.5-72B       & 0.900 & 0.899 & 0.841 & 0.836 \\
    & GPT-4o         & 0.899 & 0.897 & 0.849 & {0.842} \\
    & Gemini 2.0         & 0.893 & 0.891 & 0.843 & 0.834 \\
    \cdashline{1-6} 
    \noalign{\vskip 0.5ex}
    \multirow{5}{*}{\rotatebox{90}{w/ SC}}
    & Qwen2.5-72B       &   0.883    &    0.878   &   0.823    &   0.810   \\
    & GPT-4o         &  0.830     &   0.842    &    0.793   &   0.782    \\
    & Gemini 2.0         & 0.781 & 0.790 & 0.754 & 0.733 \\
    & SylAVL-CoT (Gemini 2.0)  & 0.765 & 0.786 & 0.730 & 0.695 \\
    & SylAVL-CoT (Gemini 2.5)  & 0.760 & 0.788 & 0.732 & 0.697 \\
    \midrule
    & & \multicolumn{4}{c}{Dubbed Lyrics $\longleftrightarrow$ Translated} \\
    \midrule
    \multirow{5}{*}{\rotatebox{90}{w/o SC}}
    & Google Translate & 0.672 & 0.649 & 0.618 & 0.600 \\
    & mBART-50       & 0.576 & 0.634 & 0.608 & 0.612 \\
    & Qwen2.5-72B       & \textbf{0.677} & \textbf{0.655} & 0.637 & 0.633 \\
    & GPT-4o         & 0.670 & 0.635 & 0.633 & 0.624 \\
    & Gemini 2.0         & 0.671 & 0.652 & 0.636 & 0.610 \\
    \cdashline{1-6} 
    \noalign{\vskip 0.5ex}
    \multirow{5}{*}{\rotatebox{90}{w/ SC}}
    & Qwen2.5-72B       &  0.673    &    0.632   &   0.645   &    0.649   \\
    & GPT-4o         &  0.672     &   0.654    &   0.647    &   {0.672}    \\
    & Gemini 2.0         & 0.654 & 0.634 & \textbf{0.651} & 0.669 \\
    & SylAVL-CoT (Gemini 2.0)  & 0.656 & 0.640 & 0.647 & 0.681 \\
    & SylAVL-CoT (Gemini 2.5)  & 0.649 & 0.641 & 0.645 & \textbf{0.688} \\
    \bottomrule
    \end{tabular}%
    }
    \caption{\textbf{Comparison with other models: Semantic scores.} The abbreviations in the table are the same as those in Table~\ref{tab:comp_syllable}.}
    \label{tab:base_metrics_mpnet_vertical}
\end{table}
\section{Experiments}
We experiment to validate the effectiveness of \textbf{\texttt{SylAVL-CoT}} and analyze the \textbf{MAVL} dataset, presenting both quantitative and qualitative results.

\subsection{Experimental Setup}
\noindent \textbf{Baseline Models. }
To compare and analyze text-based translation with our model, we selected five models capable of multilingual translation. For traditional multilingual models, we adopted Google Translate~\cite{googletranslate} and mBART-50~\cite{cho2014propertiesneuralmachinetranslation}, while for LLM models, we selected Qwen2-72B~\cite{yang2024qwen2}, GPT-4o~\cite{hurst2024:gpt}\footnote{gpt-4o-2024-08-06}, and Gemini 2.0 Flash~\cite{gemini20flash}\footnote{gemini-2.0-flash-001} with and without syllable constraints. Importantly, all baseline models were evaluated using text-only inputs (original lyrics and syllable constraints when applicable), while our \textbf{\texttt{SylAVL-CoT}} approach leverages all three modalities—text, audio, and video—simultaneously through Gemini's multimodal capabilities. Additionally, we evaluated the performance of our \textbf{\texttt{SylAVL-CoT}} approach using the more recent Gemini 2.5 Flash model to assess the impact of model evolution on translation quality.

\noindent \textbf{Evaluation Strategy. }
We conduct evaluations by comparing the translated lyrics with the original and dubbed languages. The comparison with the original language assesses how faithfully the translation preserves the essence of the source lyrics, while the comparison with the dubbed language evaluates how well the translation reflects linguistic differences in the target language.

\subsection{Comparative Analysis}
In this section, we evaluate how well the translation models incorporate the three principles described in the Section~\ref{sec:metrics}.
Additional experiments results can be found in Appendix~\ref{appendix:additional_exp}.
\begin{figure*}[t!]
    \centering
    \includegraphics[width=\linewidth]{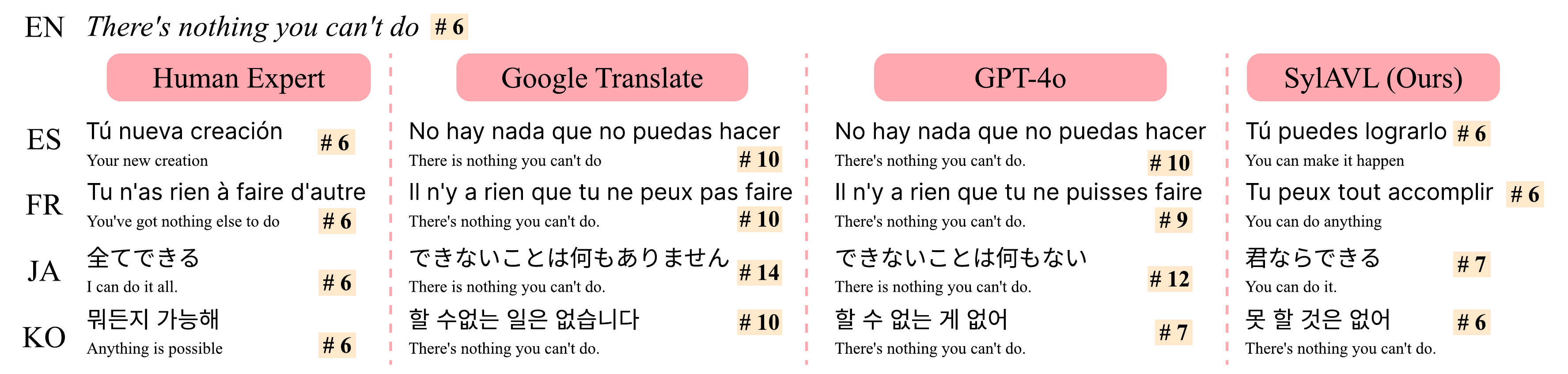}
    \caption{\textbf{Qualitative Results.} ``\#'' denotes syllable counts. The English text below each language represents its meaning. This example is ``What else can I do?'' from ``Encanto'' by Disney.}
    \label{fig:savl_qual}
\end{figure*}

\noindent \textbf{Singability. }
As shown in Table~\ref{tab:comp_syllable}, \textbf{\texttt{SylAVL-CoT}} achieves the lowest \textit{Syllable Error}, \textit{Syllable Distance}, and \textit{Error Rate} compared to both traditional machine translation models and LLM–based translation. Notably, when using the more recent Gemini 2.5 Flash model, \texttt{SylAVL-CoT} demonstrates further improvements across syllable-related metrics. This consistent improvement across languages highlights how our approach can leverage advances in underlying model capabilities.

\noindent \textbf{Sense. }
Table~\ref{tab:base_metrics_mpnet_vertical} shows that some machine translation models yield high semantic similarity scores with the original English lyrics, reflecting largely literal translations. In contrast, \textbf{\texttt{SylAVL-CoT}} shows noticeably lower similarity because it—like human experts—employs extensive paraphrasing and restructuring to maintain singability, thus deviating more from the original wording. This is shown in Figure~\ref{fig:savl_qual}.

When compared to human-translated dubbed lyrics, models with high similarity to the English original drop sharply, showing literal translations miss the creative adjustments in professional dubbing. Meanwhile, \textbf{\texttt{SylAVL-CoT}} maintains similarity scores to the dubbed lyrics that rival other models, indicating its ability to adopt strategies similar to human translators. Thus, while \textbf{\texttt{SylAVL-CoT}} sacrifices surface-level closeness to English, it preserves the deeper sense by aligning with how professionals adapt content for singability. More detailed experiments on \textit{Sense} can be found in Appendix~\ref{appendix:additional_exp}.

\begin{table}[t]
\centering
\scriptsize
\setlength{\tabcolsep}{8pt} 
\resizebox{\columnwidth}{!}{%
\begin{tabular}{p{0.5cm} l *{4}{>{\centering\arraybackslash}p{0.7cm}}} 
\toprule
\multirow{2}{*}{} & \multirow{2}{*}{Model} & \multicolumn{4}{c}{Levenshtein Distance$\downarrow$} \\
\cmidrule(lr){3-6} 
&       & ES   & FR   & KO   & JA \\
\midrule
& & \multicolumn{4}{c}{English Lyrics $\longleftrightarrow$ Translated} \\
\midrule
\multirow{1}{*}{} & Human Expert   & 23.22 & 24.41 & 25.07 & 25.40 \\[0.5ex]
\cdashline{1-6} 
\noalign{\vskip 0.5ex}
\multirow{5}{*}{\rotatebox{90}{w/o SC}}
& Google Translate & 26.25 & 26.70 & 34.29 & 35.65 \\
& MBart-50       & 50.38 & 26.83 & 35.71 & 32.13 \\
& Qwen2.5-72B       & 26.62 & 26.89 & 33.24 & 31.78 \\
& GPT-4o         & 26.92 & 27.18 & 32.29 & 32.68 \\
& Gemini 2.0         & 27.06 & 27.59 & 33.90 & 33.30 \\
\cdashline{1-6} 
\noalign{\vskip 0.5ex}
\multirow{5}{*}{\rotatebox{90}{w/ SC}}
& Qwen2.5-72B       & 25.82 & 25.81 & 29.22 & 28.60 \\
& GPT-4o         & 23.51 & \textbf{23.73} & 26.11 & 26.73 \\
& Gemini 2.0         & 24.10 & 24.97 & 26.70 & 26.63 \\
& SylAVL-CoT (Gemini 2.0)  & \underline{23.16} & \underline{24.07} & \textbf{25.20} & \textbf{25.88} \\
& SylAVL-CoT (Gemini 2.5)  & \textbf{23.14} & 24.13 & \underline{25.23} & \underline{25.90} \\
\midrule
& & \multicolumn{4}{c}{Dubbed Lyrics $\longleftrightarrow$ Translated} \\
\midrule
\multirow{5}{*}{\rotatebox{90}{w/o SC}}
& Google Translate & 20.32 & 22.32 & 30.79 & 30.29 \\
& MBart-50       & 44.89 & 23.09 & 32.63 & 26.34 \\
& Qwen2.5-72B       & 20.44 & 22.30 & 28.96 & 25.43 \\
& GPT-4o         & 20.79 & 22.81 & 28.50 & 26.15 \\
& Gemini 2.0         & 20.91 & 22.66 & 29.07 & 26.86 \\
\cdashline{1-6} 
\noalign{\vskip 0.5ex}
\multirow{5}{*}{\rotatebox{90}{w/ SC}}
& Qwen2.5-72B       & 19.49 & 21.65 & 24.65 & 21.11 \\
& GPT-4o         & 16.62 & \textbf{19.30} & 20.98 & 18.29 \\
& Gemini 2.0         & 17.86 & 21.22 & 22.14 & 18.21 \\
& SylAVL-CoT (Gemini 2.0)  & \textbf{16.25} & 20.07 & \textbf{19.98} & \underline{15.54} \\
& SylAVL-CoT (Gemini 2.5)  & \underline{16.28} & \underline{20.05} & \underline{20.04} & \textbf{15.41} \\
\bottomrule
\end{tabular}%
}
\caption{\textbf{Comparison with other models: Phonetic distance.} The abbreviations in the table are the same as those in Table~\ref{tab:comp_syllable}. \textbf{Best} and \underline{second} are highlighted.}
\label{tab:phonetic-distance}
\end{table}
\noindent \textbf{Naturalness. }
Assuming that both the original lyrics and human-translated dubbing uphold a high level of \emph{Naturalness}, we measure how closely each model's output aligns phonetically with these two references. Notably, \textbf{\texttt{SylAVL-CoT}} yields consistently lower phonetic distance scores across both references, indicating that its generated lyrics share key articulatory and rhythmic traits with professionally produced lyrics. By preserving these phonetic elements, \textbf{\texttt{SylAVL-CoT}} appears to retain a degree of naturalness similar to the original and dubbed lyrics.

\begin{table}[tbp]
\centering
\scriptsize               
\setlength{\tabcolsep}{8pt} 
\resizebox{\columnwidth}{!}{%
\begin{tabular}{l *{4}{>{\centering\arraybackslash}p{0.8cm}}}
\toprule
\multirow{2}{*}{Model} & \multicolumn{4}{c}{MPNet$\uparrow$} \\
\cmidrule(lr){2-5}
                        & ES   & FR   & KO   & JA \\
\midrule
& \multicolumn{4}{c}{English Lyrics $\longleftrightarrow$ Translated Lyrics} \\
\midrule
T                   & 0.7584 & \underline{0.7888} & \underline{0.7391} & \underline{0.6965} \\
T + V               & 0.7586 & 0.7794 & 0.7286 & 0.6952 \\
T + A               & \textbf{0.7723} & \textbf{0.8020} & \textbf{0.7484} & \textbf{0.7133} \\
T + A + V   & \underline{0.7652} & 0.7859 & 0.7298 & 0.6953 \\
\midrule
& \multicolumn{4}{c}{Dubbed Lyrics $\longleftrightarrow$ Translated Lyrics} \\
\midrule
T                   & 0.6481 & 0.6332 & 0.6398 & 0.6748 \\
T + V               & 0.6483 & 0.6284 & 0.6433 & \underline{0.6796} \\
T + A               & \underline{0.6559} & \underline{0.6400} & \underline{0.6462} & 0.6785 \\
T + A + V    & \textbf{0.6561} & \textbf{0.6402} & \textbf{0.6466} & \textbf{0.6814} \\
\bottomrule
\end{tabular}%
}
\caption{\textbf{Result of ablation study.} ``T'', ``A'', and ``V'' reprsent text, audio and video, repectively.}
\label{tab:ablation-mpnet}
\end{table}
\subsection{Ablation Studies}
Table~\ref{tab:ablation-mpnet} presents an ablation study examining the individual contributions of text, audio, and video inputs in our Multimodal CoT approach. When comparing semantic similarity with the original English lyrics, using only the audio modality yields the highest scores across all target languages.

In contrast, relying solely on video diminishes semantic similarity, suggesting that visual context prompts the model to paraphrase more freely, thus deviating further from the source text. However, when we compare outputs against the dubbed lyrics instead of the original English, combining both audio and video yields the best performance. Consequently, leveraging \textbf{all modalities} produces the most dubbing-like translations, striking an effective balance between literal accuracy and context-driven paraphrasing.

\begin{table}[tbp]
    \centering
    \scriptsize               
    \setlength{\tabcolsep}{6pt} 
    \resizebox{\columnwidth}{!}{%
    \begin{tabular}{cc *{4}{>{\centering\arraybackslash}p{0.8cm}}}
    \toprule
    \multicolumn{2}{c}{Model} & \multicolumn{4}{c}{Syllable Error$\downarrow$} \\
    \midrule
    Syllable List & Refine  & ES   & FR   & KO   & JA \\
    \midrule
    & & \multicolumn{4}{c}{English Lyrics $\longleftrightarrow$ Translated Lyrics} \\
    \midrule
    \myxmark & \myxmark                   & 3.641 & 1.860 & 3.359 & 6.069 \\
    \myxmark & \mychecking                   & 2.805 & 1.047 & 1.528 & 2.955 \\
    \mychecking & \myxmark                   & 1.353 & 0.986 & 0.879 & 3.267 \\
    \mychecking & \mychecking                   & \textbf{0.966} & \textbf{0.902} & \textbf{0.695} & \textbf{2.572} \\
    \midrule
    & & \multicolumn{4}{c}{Dubbed Lyrics $\longleftrightarrow$ Translated Lyrics} \\
    \midrule
    \myxmark & \myxmark                   & 2.740 & 2.036 & 3.244 & 5.821 \\
    \myxmark & \mychecking                   & 2.122 & \textbf{1.334} & 1.711 & 3.264 \\
    \mychecking & \myxmark                   & 1.530 & 1.337 & 1.263 & 3.683 \\
    \mychecking & \mychecking                   & \textbf{1.349} & 1.363 & \textbf{1.261} & \textbf{3.107} \\
    \bottomrule
    \end{tabular}%
    }
    \caption{\textbf{Result of ablation study.} ``\mychecking'' and ``\myxmark'' represent whether the ``Syllable List'' and ``Refining'' steps are used or not, respectively.}
    \label{tab:ablation-stage-SE}
    \end{table}
Table~\ref{tab:ablation-stage-SE} demonstrates that incorporating the ``Syllable List'' generation and ``Refine'' stages into our CoT process significantly reduces syllable errors when compared to the original English lyrics. In particular, the introduction of the ``Syllable List'' stage shows a marked improvement in reducing syllable differences across many languages compared to baselines without it. A similar trend of improvement is observed when comparing the translated lyrics against the dubbed lyrics, with the combination of both stages generally yielding the lowest \textit{Syllable Error}. The prompt used for CoT without ``Syllable List'' or ``Refine'' can be found in Appendix~\ref{appendix:cot_prompts}.

\subsection{Model Evolution Impact}
To assess how advances in pretrained models affect our approach, we compared the performance of \textbf{\texttt{SylAVL-CoT}} using different versions of Gemini. As shown in Tables~\ref{tab:comp_syllable} and \ref{tab:base_metrics_mpnet_vertical}, the transition from Gemini 2.0 Flash to 2.5 Flash yields consistent improvements. The newer model achieves better syllable constraint adherence while maintaining or slightly improving semantic similarity scores. This demonstrates that  our \textbf{\texttt{SylAVL-CoT}} methodology can effectively leverage improvements in underlying model capabilities without modification, and the MAVL benchmark serves as a valuable tool for evaluating multimodal language models' ability to handle constrained generation tasks.

\begin{table*}[!t]
\centering
\scriptsize
\setlength{\tabcolsep}{3pt}
\renewcommand{\arraystretch}{1.0}
\resizebox{\textwidth}{!}{%
\begin{tabular}{l c *{4}{>{\centering\arraybackslash}p{1.2cm}} *{4}{>{\centering\arraybackslash}p{1.2cm}} }
\toprule
\multirow{2}{*}{Model} & \multirow{2}{*}{CoT} & \multicolumn{4}{c}{Syllable Error $\downarrow$} & \multicolumn{4}{c}{MPNet Score $\uparrow$} \\
\cmidrule(lr){3-6}\cmidrule(lr){7-10}
                       &     & ES   & FR   & KO   & JA   & ES   & FR   & KO   & JA   \\
\midrule
\multicolumn{2}{c}{} & \multicolumn{8}{c}{\textbf{English Lyrics $\longleftrightarrow$ Translated Lyrics}} \\
\midrule
\multirow{2}{*}{Qwen2.5-72B}
                 & w/o CoT & 6.512 & 2.729 & 6.502 & 10.061 & \textbf{0.883} & \textbf{0.878} & \textbf{0.823} & \textbf{0.810} \\
                 & w/ CoT  & 3.440 & 1.545 & 5.785 & 4.989  & 0.834 & 0.844 & 0.805 & 0.777 \\
\midrule
\multirow{2}{*}{GPT-4o}
                 & w/o CoT & 3.164 & 1.442 & 3.084 & 7.221  & 0.830 & 0.842 & 0.793 & 0.782 \\
                 & w/ CoT  & 1.888 & 0.981 & 1.435 & 3.538  & 0.816 & 0.839 & 0.785 & 0.760 \\
\midrule
\multirow{2}{*}{Gemini 2.0}
                 & w/o CoT & 3.585 & 1.384 & 3.039 & 6.257  & 0.781 & 0.790 & 0.754 & 0.733 \\
                 & w/ CoT  & \textbf{0.976} & \textbf{0.848} & \textbf{0.717} & \textbf{2.026} & 0.758 & 0.789 & 0.739 & 0.697 \\
\midrule
\multicolumn{2}{c}{} & \multicolumn{8}{c}{\textbf{Dubbed Lyrics $\longleftrightarrow$ Translated Lyrics}} \\
\midrule
\multirow{2}{*}{Qwen2.5-72B}
                 & w/o CoT & 5.185 & 2.893 & 6.229 & 9.703  & \textbf{0.673} & 0.632 & 0.645 & 0.649 \\
                 & w/ CoT  & 2.982 & 1.792 & 5.021 & 5.944  & 0.655 & 0.652 & 0.626 & 0.654 \\
\midrule
\multirow{2}{*}{GPT-4o}
                 & w/o CoT & 2.403 & 1.723 & 3.067 & 6.940  & 0.672 & \textbf{0.654} & \textbf{0.647} & \textbf{0.672} \\
                 & w/ CoT  & 1.723 & \textbf{1.219} & 1.691 & 3.820  & 0.667 & 0.653 & 0.675 & 0.655 \\
\midrule
\multirow{2}{*}{Gemini 2.0}
                 & w/o CoT & 2.610 & 1.703 & 2.941 & 5.996  & 0.654 & 0.634 & 0.651 & 0.669 \\
                 & w/ CoT  & \textbf{1.347} & 1.297 & \textbf{1.287} & \textbf{2.637} & 0.648 & 0.633 & 0.640 & 0.675 \\
\bottomrule
\end{tabular}
}
\caption{\textbf{Impact of \texttt{SylAVL-CoT} on different LLMs.} The performance of Qwen2.5-72B, GPT-4o, and Gemini with and without \textbf{\texttt{SylAVL-CoT}} prompting (text-only, no multimodal inputs).}
\label{tab:cot_comparison}
\end{table*}

Furthermore, we investigated whether Gemini performs well with our CoT approach compared to other LLMs. As shown in Table~\ref{tab:cot_comparison}, we compared Qwen2.5-72B, GPT-4o, and Gemini with and without CoT prompting using text-only inputs to isolate the effect of CoT reasoning. While GPT-4o achieve similar syllable constraint adherence in their non-CoT configurations, Gemini demonstrates dramatically superior performance when using CoT.

\begin{table}[!htb]
\centering
\scriptsize
\setlength{\tabcolsep}{4pt} 
\resizebox{\columnwidth}{!}{%
\begin{tabular}{llccc}
    \toprule
    \textbf{Language} & \textbf{Model} & \textbf{Singability} & \textbf{Sense} & \textbf{Overall Quality} \\
    \midrule
    \multirow{6}{*}{Spanish} 
    & Human Expert & 3.90{\tiny\textpm 1.11} & 3.46{\tiny\textpm 1.20} & 3.61{\tiny\textpm 1.15} \\
    \cdashline{2-5}
    \noalign{\vskip 0.5ex}
    & Google Translate & 2.12{\tiny\textpm 1.27} & 2.60{\tiny\textpm 1.36} & 2.18{\tiny\textpm 1.25} \\
    & Qwen2.5-72B & 2.94{\tiny\textpm 1.24} & \textbf{4.05}{\tiny\textpm 1.00} & 3.26{\tiny\textpm 1.13} \\
    & GPT-4 & 3.25{\tiny\textpm 1.18} & 3.90{\tiny\textpm 1.15} & 3.43{\tiny\textpm 1.13} \\
    & Gemini 2.0 & 3.16{\tiny\textpm 1.06} & 3.22{\tiny\textpm 1.06} & 3.03{\tiny\textpm 0.98} \\
    & SylAVL-CoT & \textbf{3.68}{\tiny\textpm 1.16} & 3.46{\tiny\textpm 1.34} & \textbf{3.57}{\tiny\textpm 1.26} \\
    \midrule
    \multirow{6}{*}{French} 
    & Human Expert & 3.94{\tiny\textpm 1.23} & 3.55{\tiny\textpm 1.17} & 3.50{\tiny\textpm 1.23} \\
    \cdashline{2-5}
    \noalign{\vskip 0.5ex}
    & Google Translate & 3.39{\tiny\textpm 1.17} & 3.80{\tiny\textpm 1.05} & 3.57{\tiny\textpm 1.32} \\
    & Qwen2.5-72B & 3.85{\tiny\textpm 1.17} & \textbf{3.89}{\tiny\textpm 0.95} & 3.85{\tiny\textpm 1.23} \\
    & GPT-4 & 3.86{\tiny\textpm 1.05} & 3.86{\tiny\textpm 1.01} & 3.86{\tiny\textpm 1.14} \\
    & Gemini 2.0 & 3.86{\tiny\textpm 1.18} & 3.65{\tiny\textpm 1.07} & 3.70{\tiny\textpm 1.19} \\
    & SylAVL-CoT & \textbf{4.04}{\tiny\textpm 1.03} & 3.86{\tiny\textpm 0.98} & \textbf{3.93}{\tiny\textpm 1.11} \\
    \midrule
    \multirow{6}{*}{Korean} 
    & Human Expert & 3.88{\tiny\textpm 1.24} & 2.94{\tiny\textpm 1.44} & 3.28{\tiny\textpm 1.28} \\
    \cdashline{2-5}
    \noalign{\vskip 0.5ex}
    & Google Translate & 2.05{\tiny\textpm 1.23} & 2.54{\tiny\textpm 1.39} & 2.15{\tiny\textpm 1.21} \\
    & Qwen2.5-72B & 2.00{\tiny\textpm 1.34} & 3.36{\tiny\textpm 1.12} & 2.46{\tiny\textpm 1.04} \\
    & GPT-4 & 3.41{\tiny\textpm 1.10} & 3.33{\tiny\textpm 1.23} & 3.19{\tiny\textpm 1.07} \\
    & Gemini 2.0 & 3.47{\tiny\textpm 1.22} & 3.29{\tiny\textpm 1.41} & 3.25{\tiny\textpm 1.26} \\
    & SylAVL-CoT & \textbf{4.32}{\tiny\textpm 0.81} & \textbf{3.71}{\tiny\textpm 1.24} & \textbf{3.95}{\tiny\textpm 1.08} \\
    \midrule
    \multirow{6}{*}{Japanese} 
    & Human Expert & 3.89{\tiny\textpm 0.80} & 3.43{\tiny\textpm 0.96} & 3.57{\tiny\textpm 1.02} \\
    \cdashline{2-5}
    \noalign{\vskip 0.5ex}
    & Google Translate & 2.21{\tiny\textpm 1.11} & 2.56{\tiny\textpm 1.13} & 2.28{\tiny\textpm 1.13} \\
    & Qwen2.5-72B & 3.17{\tiny\textpm 1.01} & 3.31{\tiny\textpm 0.81} & 3.19{\tiny\textpm 0.96} \\
    & GPT-4 & 3.15{\tiny\textpm 0.98} & 3.39{\tiny\textpm 0.76} & 3.19{\tiny\textpm 0.98} \\
    & Gemini 2.0 & 3.36{\tiny\textpm 1.01} & \textbf{3.65}{\tiny\textpm 0.77} & 3.33{\tiny\textpm 1.04} \\
    & SylAVL-CoT & \textbf{3.84}{\tiny\textpm 0.79} & 3.60{\tiny\textpm 0.88} & \textbf{3.64}{\tiny\textpm 0.84} \\
    \bottomrule
\end{tabular}
}
\caption{\textbf{Result of user study.} Mean scores (\textpm standard deviation) from native speakers ($N=10$ per language) evaluating lyric translations by various models (all syllable-constrained except Google Translate). Metrics include \textit{Singability}, \textit{Sense}, and \textit{Overall Quality}. See Appendix~\ref{appendix:userstudy-detail} for details.}
\label{tab:user_study_results}
\end{table}

This disparity reveals fundamental differences in how these models process complex, multi-constraint tasks. Qwen2.5 and GPT-4o appear optimized for direct translation but struggle to maintain syllable constraints when reasoning through steps. In contrast, Gemini excels at following structured reasoning paths that involve tracking multiple constraints simultaneously. These findings validate our choice of Gemini for \textbf{\texttt{SylAVL-CoT}} and demonstrate that successful multimodal lyrics translation requires not just language understanding, but also the ability to maintain complex constraints throughout a reasoning process.

\subsection{User Study}
We conducted a user study to evaluate lyric translations generated by various models. All models evaluated, except for Google Translate, are syllable-constrained. More details regarding our user study methodology, including participant recruitment and task design, can be found in Appendix~\ref{appendix:userstudy-detail}.

Results in the Table~\ref{tab:user_study_results} indicate that \textbf{\texttt{SylAVL-CoT}} achieved higher \textit{Overall Quality} scores than other models across all languages. Looking at specific aspects, \textbf{\texttt{SylAVL-CoT}} demonstrated notably high scores in \textit{Singability} for most languages. However, in French, \textbf{\texttt{SylAVL-CoT}}'s advantage in \textit{Singability} over other syllable-constrained models was relatively smaller. This observation aligns with Table \ref{tab:comp_syllable}, which reportedly shows that other models already exhibit lower \textit{Syllable Error} in French compared to other languages, making \textbf{\texttt{SylAVL-CoT}}'s lead less pronounced. Furthermore, regarding \textit{Sense}, \textbf{\texttt{SylAVL-CoT}} managed to maintain scores comparable to, and sometimes better than, other models, even while prioritizing \textit{Singability}. This balance ultimately contributed to its superior \textit{Overall Quality}.

\section{Conclusion}
In this paper, we introduced Multilingual Audio-Video Lyrics Benchmark (\textbf{MAVL}), the first multilingual and multimodal parallel lyrics translation benchmark that integrates text, audio, and video for singable translations. We also proposed \textbf{\texttt{SylAVL-CoT}}, which leverages existing MLLMs without fine-tuning and enforces syllable constraints through Chain-of-Thought reasoning. Our experiments show that \textbf{\texttt{SylAVL-CoT}} balances expressive paraphrasing with contextual accuracy, addressing a key gap in musical animations. We hope these advances pave the way for new automated lyrics translation systems and further research in multilingual, multimodal machine translation.
\section{Limitations}
\label{sec:limitations}

While the proposed \textbf{MAVL} dataset and the \textbf{\texttt{SylAVL-CoT}} framework address several challenges in multilingual, multimodal lyrics translation, there are still limitations:

\paragraph{Data Scope.}
Our dataset mainly focuses on animated musicals and on five target languages (Spanish, French, Japanese, and Korean, in addition to the original English). Although the dataset provides a rich testing ground for multimodal translation, their thematic variety may not represent the full range of musical genres, languages, and styles encountered in broader contexts. Consequently, models tested solely on this data may not generalize well to other genre-specific lyrical structures, or under-represented low-resource languages. 

However, we view MAVL as a foundational step toward broader coverage. The dataset collection pipeline and quality standards established here can serve as a blueprint for expanding to other genres. For instance, the insights gained from MAVL could be leveraged to develop automated quality assessment models that identify high-quality translations in genres like pop music or K-pop, where fan-made translations are abundant but vary in quality. Additionally, our framework could be extended to semi-automatically curate datasets for low-resource languages by combining professional translations where available with carefully filtered community contributions.

\paragraph{Line-based translation.}
Our current \textbf{\texttt{SylAVL-CoT}} approach primarily translates lyrics on a line-by-line basis. However, effective lyric translation often benefits from more flexible strategies, such as rephrasing across line breaks, or splitting and merging lines, which can significantly enhance singability and poetic expression. Such holistic strategies are often best implemented at a section level rather than a strict line level, allowing for more natural rhythmic and semantic flow. While the \textbf{MAVL} dataset includes section-level annotations, our current model does not fully leverage this. We anticipate that future research could utilize these section-level annotations to explore more sophisticated, context-aware translation strategies that transcend single-line processing, leading to more natural and musically-fitting translations.

\paragraph{Tonal Language Application.}
Our framework, aiming for broad multilingual applicability, does not currently incorporate specialized mechanisms for tonal languages such as Chinese. As highlighted in \citep{guo2022automaticsongtranslationtonal} and \citep{ye2024singitnarrateit}, translating lyrics into tonal languages often requires specific considerations for tone contours to preserve musicality and meaning, which can involve distinct processing steps. Our pursuit of a general-purpose solution meant these language-specific tonal constraints were not a primary focus. We hope that future iterations of our work can be extended to address the unique challenges of tonal languages, potentially by integrating or adapting techniques from existing research to enhance performance in these linguistic contexts.

\paragraph{Alignment Challenges.}
Precise synchronization of lyrics, audio, and video in musical settings remains non-trivial. Although we employ techniques such as Whisper-based alignment and careful human annotation, discrepancies can persist, especially for lines containing overlapping voices, spoken dialogue, or onomatopoeic interjections. These alignment inaccuracies may lead to sub-optimal multimodal model training or evaluation. Future work could incorporate more robust audio-visual alignment methods or user-in-the-loop correction to refine time stamping for each lyric segment.

\paragraph{Evaluation Metrics.}
Although our evaluation framework focuses on singability, sense, naturalness, these metrics still cannot completely capture musicality or artistic style. Automated metrics do not fully reflect subjective audience judgments. Additionally, cultural references and emotional nuance might be lost in translation and remain difficult to quantify objectively. Future work could explore LLM-based evaluation frameworks to better capture these subjective aspects, potentially using multi-agent systems where different agents assess rhythmic fit, emotional alignment, and poetic quality.

\paragraph{Broader Applicability.}
Our emphasis on animated musical translations may not translate directly to other domains such as live theater, opera, pop music, or user-generated musical content. The complexities in live performances, spontaneous improvisations, or multi-speaker settings are beyond the current benchmark's scope. Future research could extend the approach to a wider range of musical and performance contexts to validate the model's robustness and adaptability.

\smallskip
Overall, these limitations highlight the need for broader, more diversified datasets, refined alignment techniques, and more holistic metrics to capture the creative and performative aspects of lyric translation. We hope that releasing \textbf{MAVL} and proposing \textbf{\texttt{SylAVL-CoT}} spark further innovations and encourage the research community to build on or address these limitations in future work.

\section{Ethical Consideration}
In conducting this research, we have taken various ethical aspects into account to ensure responsible and fair practices in the development and dissemination of our work.

\noindent \paragraph{Transparency. } We are committed to maintaining transparency throughout our research process. All preprocessing steps, alignment techniques, and model training methodologies are fully disclosed to enable reproducibility and facilitate further scrutiny by the research community. More details are in Appendix.

\noindent \paragraph{Copyright Compliance. } To respect intellectual property rights, our dataset does not distribute copyrighted materials directly. Instead, we provide structured metadata and download links where applicable, ensuring compliance with copyright regulations while preserving the dataset's usability for research.

\noindent \paragraph{Cultural Sensitivity and Inclusivity. } Our research aims to contribute to a diverse and inclusive representation of musical content across languages. We have taken measures to respect cultural nuances and avoid biases, ensuring that our approach promotes fairness in singable lyric translation.

\noindent \paragraph{Potential Societal Impact. } We acknowledge that our work may have broader implications for cross-cultural communication and creative industries. We encourage continued ethical reflection on how AI-assisted lyric translation can be leveraged responsibly, particularly in artistic and commercial applications.

\section*{Acknowledgements}
This work was supported by the Institute of Information \& Communications Technology Planning \& Evaluation (IITP) grants funded by the Korean government (MSIT) (No. RS-2024-00457882, AI Research Hub Project) and by the National Research Foundation of Korea (NRF) grants funded by the Korean government (MSIT) (Nos. RS-2024-00354218 and RS-2024-00353125).

\bibliography{custom}

@misc{kim2024kpoplyrictranslationdataset,
      title={K-pop Lyric Translation: Dataset, Analysis, and Neural-Modelling}, 
      author={Haven Kim and Jongmin Jung and Dasaem Jeong and Juhan Nam},
      year={2024},
      eprint={2309.11093},
      archivePrefix={arXiv},
      primaryClass={cs.CL},
      url={https://arxiv.org/abs/2309.11093}, 
}

@inproceedings{guo2022automaticsongtranslationtonal,
  title={Automatic Song Translation for Tonal Languages},
  author={Guo, Fenfei and Zhang, Chen and Zhang, Zhirui and He, Qixin and Zhang, Kejun and Xie, Jun and Boyd-Graber, Jordan},
  booktitle={Findings of the Association for Computational Linguistics: ACL 2022},
  pages={729--743},
  year={2022}
}

@article{Leni2019ANALYZINGTS,
  title={ANALYZING TRANSLATION STRATEGIES UTILIZED IN THE TRANSLATION OF SONG “DO YOU WANT TO BUILD A SNOWMAN?”},
  author={Chrisna Leni and Athriyana Santye Pattiwael},
  journal={Journal of Language and Literature},
  year={2019},
  url={https://api.semanticscholar.org/CorpusID:187067408}
}

@misc{ou2023songsborderssingablecontrollable,
      title={Songs Across Borders: Singable and Controllable Neural Lyric Translation}, 
      author={Longshen Ou and Xichu Ma and Min-Yen Kan and Ye Wang},
      year={2023},
      eprint={2305.16816},
      archivePrefix={arXiv},
      primaryClass={cs.CL},
      url={https://arxiv.org/abs/2305.16816}, 
}

@misc{li2023translatebeautysongsjointly,
      title={Translate the Beauty in Songs: Jointly Learning to Align Melody and Translate Lyrics}, 
      author={Chengxi Li and Kai Fan and Jiajun Bu and Boxing Chen and Zhongqiang Huang and Zhi Yu},
      year={2023},
      eprint={2303.15705},
      archivePrefix={arXiv},
      primaryClass={cs.CL},
      url={https://arxiv.org/abs/2303.15705}, 
}

@misc{kim2023computationalevaluationframeworksingable,
      title={A Computational Evaluation Framework for Singable Lyric Translation}, 
      author={Haven Kim and Kento Watanabe and Masataka Goto and Juhan Nam},
      year={2023},
      eprint={2308.13715},
      archivePrefix={arXiv},
      primaryClass={cs.CL},
      url={https://arxiv.org/abs/2308.13715}, 
}

@misc{radford2022robustspeechrecognitionlargescale,
      title={Robust Speech Recognition via Large-Scale Weak Supervision}, 
      author={Alec Radford and Jong Wook Kim and Tao Xu and Greg Brockman and Christine McLeavey and Ilya Sutskever},
      year={2022},
      eprint={2212.04356},
      archivePrefix={arXiv},
      primaryClass={eess.AS},
      url={https://arxiv.org/abs/2212.04356}, 
}

@incollection{low2005pentathlon,
  title={The pentathlon approach to translating songs},
  author={Low, Peter},
  booktitle={Song and significance},
  pages={185--212},
  year={2005},
  publisher={Brill}
}

@article{franzon2008choices,
  title={Choices in song translation: Singability in print, subtitles and sung performance},
  author={Franzon, Johan},
  journal={The Translator},
  volume={14},
  number={2},
  pages={373--399},
  year={2008},
  publisher={Taylor \& Francis}
}

@inproceedings{ye2024singitnarrateit,
  title={Sing it, Narrate it: Quality Musical Lyrics Translation},
  author={Ye, Zhuorui and Li, Jinhan and Xu, Rongwu},
  booktitle={Findings of the Association for Computational Linguistics: EMNLP 2024},
  pages={5498--5520},
  year={2024}
}

@misc{ma2025audiocotexploringchainofthoughtreasoning,
      title={Audio-CoT: Exploring Chain-of-Thought Reasoning in Large Audio Language Model}, 
      author={Ziyang Ma and Zhuo Chen and Yuping Wang and Eng Siong Chng and Xie Chen},
      year={2025},
      eprint={2501.07246},
      archivePrefix={arXiv},
      primaryClass={cs.SD},
      url={https://arxiv.org/abs/2501.07246}, 
}

@misc{zhang2024multimodalchainofthoughtreasoninglanguage,
      title={Multimodal Chain-of-Thought Reasoning in Language Models}, 
      author={Zhuosheng Zhang and Aston Zhang and Mu Li and Hai Zhao and George Karypis and Alex Smola},
      year={2024},
      eprint={2302.00923},
      archivePrefix={arXiv},
      primaryClass={cs.CL},
      url={https://arxiv.org/abs/2302.00923}, 
}

@misc{hu2024largelanguagemodelmultidomain,
      title={Large Language Model for Multi-Domain Translation: Benchmarking and Domain CoT Fine-tuning}, 
      author={Tianxiang Hu and Pei Zhang and Baosong Yang and Jun Xie and Derek F. Wong and Rui Wang},
      year={2024},
      eprint={2410.02631},
      archivePrefix={arXiv},
      primaryClass={cs.CL},
      url={https://arxiv.org/abs/2410.02631}, 
}

@misc{du2024cotstenhancingllmbasedspeech,
      title={CoT-ST: Enhancing LLM-based Speech Translation with Multimodal Chain-of-Thought}, 
      author={Yexing Du and Ziyang Ma and Yifan Yang and Keqi Deng and Xie Chen and Bo Yang and Yang Xiang and Ming Liu and Bing Qin},
      year={2024},
      eprint={2409.19510},
      archivePrefix={arXiv},
      primaryClass={cs.CL},
      url={https://arxiv.org/abs/2409.19510}, 
}

@article{1075843ar,
  title="A multimodal model of analysis for the translation of songs from stage musicals",
  author="Carpi, Beatrice",
  journal="Meta",
  volume="65",
  number="2",
  pages="420--439",
  year="2020",
  publisher="Les Presses de l’Université de Montréal",
  doi="https://doi.org/10.7202/1075843ar"
}

@inproceedings{Pidhrushna2021FunctionalAT,
  title={Functional approach to songs in film translation: Challenges and compromises},
  author={Olena Pidhrushna},
  year={2021},
  url={https://api.semanticscholar.org/CorpusID:236648672}
}

@article{low2003singable,
  title={Singable translations of songs},
  author={Low, Peter},
  journal={Perspectives: Studies in Translatology},
  volume={11},
  number={2},
  pages={87--103},
  year={2003},
  publisher={Taylor \& Francis}
}

@article{susam2008translation,
author = {Susam-Sarajeva, Şebnem},
year = {2008},
month = {02},
pages = {187-200},
title = {Translation and Music},
volume = {14},
journal = {The Translator},
doi = {10.1080/13556509.2008.10799255}
}

@article{taylor2016multimodal,
author = {Taylor, Christopher},
year = {2016},
month = {08},
pages = {},
title = {The multimodal approach in audiovisual translation},
volume = {28},
journal = {Target},
doi = {10.1075/target.28.2.04tay}
}

@article{supardi2018audio,
author = {Supardi, Moh and Putri, Dea},
year = {2018},
month = {10},
pages = {399-414},
title = {Audio-Visual Translation Techniques: Subtitling and Dubbing of Movie Soundtrack in Frozen: Let it Go},
volume = {24},
journal = {Buletin Al-Turas},
doi = {10.15408/bat.v24i2.8621}
}

@article{banos2009prefabricated,
  title={Prefabricated orality: A challenge in audiovisual translation},
  author={Ba{\~n}os Pi{\~n}ero, Roc{\'\i}o and Chaume, Frederic},
  year={2009},
  publisher={Universit{\`a} di Bologna}
}

@inproceedings{rajpoot2024:multimodal,
  title={Multimodal Machine Translation for Low-Resource Indic Languages: A Chain-of-Thought Approach Using Large Language Models},
  author={Rajpoot, Pawan and Bhat, Nagaraj and Shrivastava, Ashish},
  booktitle={Proceedings of the Ninth Conference on Machine Translation},
  pages={833--838},
  year={2024}
}

@inproceedings{he2024:multi,
  title={Multi-modal latent space learning for chain-of-thought reasoning in language models},
  author={He, Liqi and Li, Zuchao and Cai, Xiantao and Wang, Ping},
  booktitle={Proceedings of the AAAI Conference on Artificial Intelligence},
  volume={38},
  number={16},
  pages={18180--18187},
  year={2024}
}

@article{wei2022:chain,
  title={Chain-of-thought prompting elicits reasoning in large language models},
  author={Wei, Jason and Wang, Xuezhi and Schuurmans, Dale and Bosma, Maarten and Xia, Fei and Chi, Ed and Le, Quoc V and Zhou, Denny and others},
  journal={Advances in neural information processing systems},
  volume={35},
  pages={24824--24837},
  year={2022}
}

@inproceedings{wang2024:videocot,
  title={VideoCoT: A Video Chain-of-Thought Dataset with Active Annotation Tool},
  author={Wang, Yan and Zeng, Yawen and Zheng, Jingsheng and Xing, Xiaofen and Xu, Jin and Xu, Xiangmin},
  booktitle={Proceedings of the 3rd Workshop on Advances in Language and Vision Research (ALVR)},
  pages={92--101},
  year={2024}
}

@inproceedings{qian2024:large,
  title={What do Large Language Models Need for Machine Translation Evaluation?},
  author={Qian, Shenbin and Sindhujan, Archchana and Kabra, Minnie and Kanojia, Diptesh and Ora{\v{s}}an, Constantin and Ranasinghe, Tharindu and Blain, Fred},
  booktitle={Proceedings of the 2024 Conference on Empirical Methods in Natural Language Processing},
  pages={3660--3674},
  year={2024}
}

@inproceedings{barrault2018:findings,
  title={Findings of the third shared task on multimodal machine translation},
  author={Barrault, Lo{\"\i}c and Bougares, Fethi and Specia, Lucia and Lala, Chiraag and Elliott, Desmond and Frank, Stella},
  booktitle={Third Conference on Machine Translation (WMT18)},
  volume={2},
  pages={308--327},
  year={2018}
}

@article{hu2024:chain,
  title={Chain-of-Thought Prompting for Speech Translation},
  author={Hu, Ke and Chen, Zhehuai and Yang, Chao-Han Huck and {\.Z}elasko, Piotr and Hrinchuk, Oleksii and Lavrukhin, Vitaly and Balam, Jagadeesh and Ginsburg, Boris},
  journal={arXiv preprint arXiv:2409.11538},
  year={2024}
}

@misc{cho2014propertiesneuralmachinetranslation,
      title={On the Properties of Neural Machine Translation: Encoder-Decoder Approaches}, 
      author={Kyunghyun Cho and Bart van Merrienboer and Dzmitry Bahdanau and Yoshua Bengio},
      year={2014},
      eprint={1409.1259},
      archivePrefix={arXiv},
      primaryClass={cs.CL},
      url={https://arxiv.org/abs/1409.1259}, 
}

@article{yang2024qwen2,
  title={Qwen2. 5 technical report},
  author={Yang, An and Yang, Baosong and Zhang, Beichen and Hui, Binyuan and Zheng, Bo and Yu, Bowen and Li, Chengyuan and Liu, Dayiheng and Huang, Fei and Wei, Haoran and others},
  journal={arXiv preprint arXiv:2412.15115},
  year={2024}
}

@inproceedings{snover-etal-2006-study,
    title = "A Study of Translation Edit Rate with Targeted Human Annotation",
    author = "Snover, Matthew  and
      Dorr, Bonnie  and
      Schwartz, Rich  and
      Micciulla, Linnea  and
      Makhoul, John",
    booktitle = "Proceedings of the 7th Conference of the Association for Machine Translation in the Americas: Technical Papers",
    month = aug # " 8-12",
    year = "2006",
    address = "Cambridge, Massachusetts, USA",
    publisher = "Association for Machine Translation in the Americas",
    url = "https://aclanthology.org/2006.amta-papers.25/",
    pages = "223--231"
}

@inproceedings{papineni2002:bleu,
  title={Bleu: a method for automatic evaluation of machine translation},
  author={Papineni, Kishore and Roukos, Salim and Ward, Todd and Zhu, Wei-Jing},
  booktitle={Proceedings of the 40th annual meeting of the Association for Computational Linguistics},
  pages={311--318},
  year={2002}
}

@article{laver1994:principles,
  title={Principles of phonetics},
  author={Laver, John},
  journal={Cambridge: Cambridge Univ},
  year={1994}
}

@article{levenshtein1966:binary,
  title={Binary codes capable of correcting deletions, insertions, and reversals},
  author={Levenshtein, VI},
  journal={Proceedings of the Soviet physics doklady},
  year={1966}
}

@misc{song2020mpnetmaskedpermutedpretraining,
      title={MPNet: Masked and Permuted Pre-training for Language Understanding}, 
      author={Kaitao Song and Xu Tan and Tao Qin and Jianfeng Lu and Tie-Yan Liu},
      year={2020},
      eprint={2004.09297},
      archivePrefix={arXiv},
      primaryClass={cs.CL},
      url={https://arxiv.org/abs/2004.09297}, 
}

@misc{gemini20flash,
  author    = {Google DeepMind},
  title     = {Gemini 2.0 Flash},
  year      = {2024},
  url       = {https://deepmind.google/technologies/gemini/},
  note      = {Accessed: 2025-02-15}
}

@misc{googletranslate,
  author    = {Google},
  title     = {Google Translate},
  url       = {https://translate.google.com/},
  note      = {Accessed: 2025-02-15}
}

@article{hurst2024:gpt,
  title={Gpt-4o system card},
  author={Hurst, Aaron and Lerer, Adam and Goucher, Adam P and Perelman, Adam and Ramesh, Aditya and Clark, Aidan and Ostrow, AJ and Welihinda, Akila and Hayes, Alan and Radford, Alec and others},
  journal={arXiv preprint arXiv:2410.21276},
  year={2024}
}

@InProceedings{Mortensen-et-al:2018,
  author = {Mortensen, David R.  and Dalmia, Siddharth and Littell, Patrick},
  title = {Epitran: Precision {G2P} for Many Languages},
  booktitle = {Proceedings of the Eleventh International Conference on Language Resources and Evaluation (LREC 2018)},
  year = {2018},
  month = {May},
  date = {7--12},
  location = {Miyazaki, Japan},
  editor = {Nicoletta Calzolari (Conference chair) and Khalid Choukri and Christopher Cieri and Thierry Declerck and Sara Goggi and Koiti Hasida and Hitoshi Isahara and Bente Maegaard and Joseph Mariani and H\'el\`ene Mazo and Asuncion Moreno and Jan Odijk and Stelios Piperidis and Takenobu Tokunaga},
  publisher = {European Language Resources Association (ELRA)},
  address = {Paris, France},
  isbn = {979-10-95546-00-9},
  language = {english}
  }

@article{Xu2024LLaVACoTLV,
  title={LLaVA-CoT: Let Vision Language Models Reason Step-by-Step},
  author={Guowei Xu and Peng Jin and Hao Li and Yibing Song and Lichao Sun and Li Yuan},
  journal={ArXiv},
  year={2024},
  volume={abs/2411.10440},
  url={https://api.semanticscholar.org/CorpusID:274116688}
}

\clearpage

\appendix
\section{Explanations of the choice for the Metrics}
\label{appendix:metrics_selection}

Our evaluation approach enhances traditional methods by incorporating comparisons not only with original lyrics but also with actual dubbed translated lyrics. This dual comparison better reflects cultural and linguistic nuances and critically assesses whether the translated lyrics can be articulated naturally in the target language. The rationale for selecting each evaluation metric is detailed below.

\subsection{Evaluation for ``Singability'' (Syllable-focused)}
We prioritized syllable-based measurements for singability as they more directly reflect the difficulty of singing compared to phoneme-level analyses. This approach is supported by previous studies~\cite{guo2022automaticsongtranslationtonal, ye2024singitnarrateit, kim2024kpoplyrictranslationdataset}, which have demonstrated the effectiveness of syllable-level analysis in assessing singing performance.

\paragraph{Syllable Error (SE)}
As noted by \cite{ye2024singitnarrateit}, an increase in the number of syllables to be pronounced generally makes singing more challenging. To capture this, we employed a weighted method using the parameter $\beta$.
The choice of $\beta$ is crucial for accurately reflecting singing difficulty. Following \citep{ye2024singitnarrateit}, we set $\beta = 2$. This specific value models the principle that singing difficulty increases more than linearly (proportionally, in this context implying a significant impact) with an increasing number of syllables. A higher $\beta$ penalizes excessive syllables more heavily, which aligns with the practical observation that cramming too many syllables into a musical phrase significantly degrades singability. Thus, $\beta=2$ is a critical setting for evaluating how well the translated lyrics maintain a singable syllable count.

\paragraph{Syllable Count Distance (SCD)}
Proposed by \citep{kim2024kpoplyrictranslationdataset}, SCD measures the congruity between the translated lyrics and the original structure. It achieves this by calculating a normalized relative distance based on the absolute differences in syllable counts, considering both the original-to-translation and translation-to-original directions. This metric is valuable for assessing not only translation accuracy in terms of length but also the consistency of rhythmic structure between the source and target lyrics.

\paragraph{Error Rate}
This metric provides a straightforward measure of singability by quantifying the proportion of incorrect syllables relative to the correct or reference values. Its simplicity offers an intuitive way to evaluate overall singability and helps in identifying common error patterns in syllable mapping.

\subsection{Evaluation for ``Sense''}
We opted for deep learning-based methods to evaluate ``Sense'' because traditional metrics like BLEU and METEOR, which rely on n-gram overlap or word matching, are often inadequate for lyric translation. Lyrics frequently require creative adaptation rather than literal translation to preserve the song's intended meaning, emotional impact, and artistic essence. Deep learning models are better equipped to assess these nuanced translations by considering contextual meaning.

\paragraph{MPNet-based semantic score}
To evaluate semantic accuracy across multiple languages, we utilized state-of-the-art deep learning models. Specifically, we employed multilingual sentence transformers from the SBERT library, which represent the current leading technology for assessing multilingual semantic similarity. This allows for a more robust evaluation of whether the core meaning of the lyrics is preserved post-translation. We also explored alternative metrics, the results of which are detailed in Table~\ref{tab:abl_combined_metrics}.

\subsection{Evaluation for ``Naturalness''}
The naturalness of translated lyrics is paramount for their acceptance and performance. We assess this through phonetic similarity.

\paragraph{Phonetic Distance}
To quantify ``Naturalness,'' we measure the phonetic similarity using Levenshtein distance calculated on the International Phonetic Alphabet (IPA) transcriptions of the translated lyrics and the reference lyrics (either original or dubbed). This serves as an effective proxy for naturalness for two main reasons:
\begin{enumerate}
    \item It quantifies the phonetic deviation from reference lyrics, which are assumed to be inherently natural and pronounceable in their respective languages. A lower distance suggests that the translation inherits this natural phonetic structure, making it more likely to sound fluent.
    \item Higher phonetic similarity (i.e., lower Levenshtein distance) implies that the translated lyrics are easier to pronounce and possess a smoother phonetic flow, mirroring the articulatory ease of the reference lyrics. This contributes significantly to the perceived naturalness of the translation when sung.
\end{enumerate}

\section{Analysis on the Dataset}
\subsection{Similarity Distribution across Languages}
\begin{figure}[htb!]
    \centering
    \includegraphics[width=\linewidth]{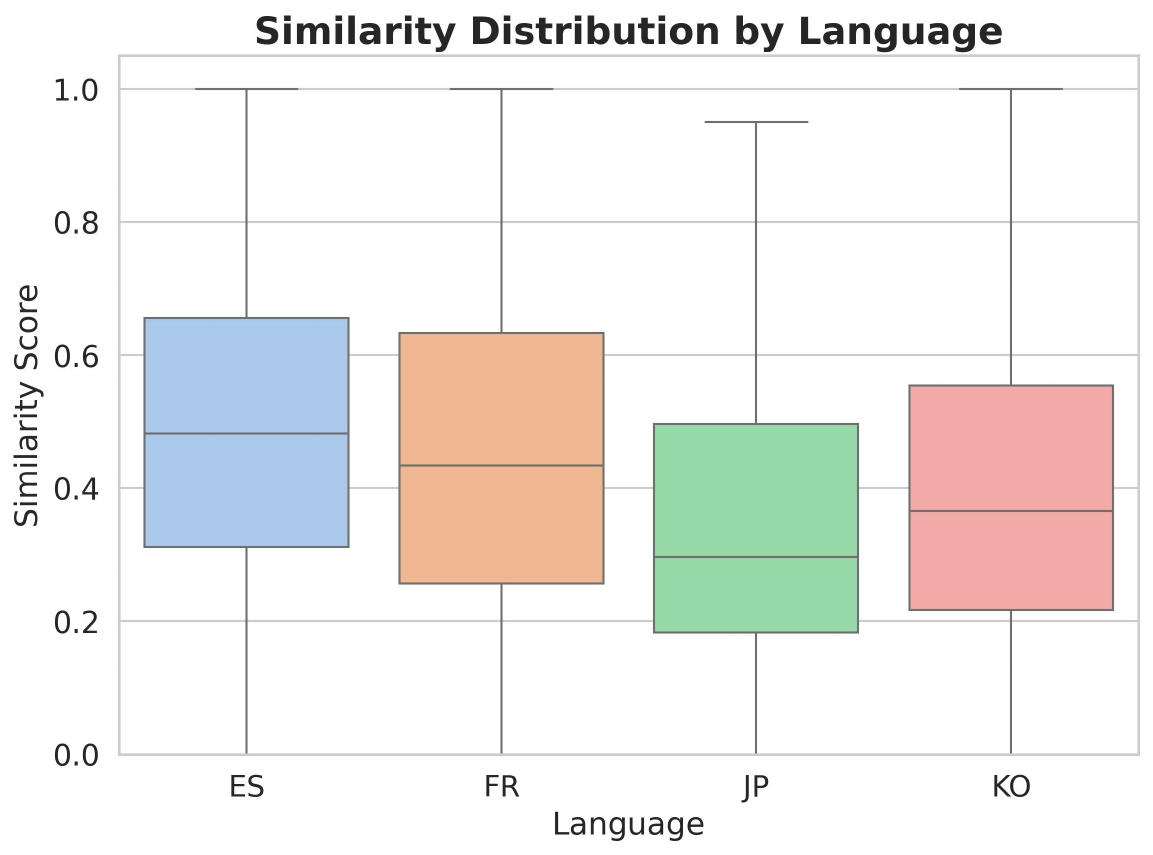}
    \caption{\textbf{MPNet Similarity Distribution By Languages.}}
    \label{appendix:similarity_distribution}
\end{figure}
Figure~\ref{appendix:similarity_distribution} illustrates the MPNet similarity distribution between the original English lyrics and their dubbed counterparts across various languages. It is important to note the methodology used for this specific visualization. Unlike the multilingual MPNet approach potentially discussed elsewhere, here we aimed to mitigate biases arising from inherent linguistic distances. Such distances could skew similarity scores even when translations are relatively literal, potentially misrepresenting the true degree of translational fidelity. Therefore, the dubbed lyrics for each language (ES, FR, JP, KR) were first translated into English using Google Translate. Subsequently, the similarity between these English-translated dubbed lyrics and the original English lyrics was computed using an English-specific MPNet\footnote{https://huggingface.co/sentence-transformers/all-mpnet-base-v2}. This approach was chosen to enhance the reliability of the analysis; however, it is worth noting that when directly measuring the similarity between the original English lyrics and the dubbed lyrics using a multilingual MPNet, the results were consistent with the findings presented here.
The chart reveals distinct similarity patterns, with scores generally following the order: ES, FR, KO, and then JA. This observation suggests that languages like Spanish (ES) and French (FR), which share more structural and alphabetical similarities with English, tend to feature more literal translations, resulting in higher similarity scores. Conversely, East Asian languages, particularly Japanese (JA), exhibit lower similarity. This indicates that their dubbed versions likely contain more significantly different expressions or a greater degree of free translation, reflecting the substantial linguistic divergence from English.  These findings are also corroborated by the Human Expert Row in Table~\ref{tab:base_metrics_mpnet_vertical}.

\begin{table*}[htbp]
    \centering
    \scriptsize               
    \setlength{\tabcolsep}{4pt} 
    \resizebox{\textwidth}{!}{%
    \begin{tabular}{c c *{4}{c} | *{4}{c} | *{4}{c}}
    \toprule
    \multirow{2}{*}{} & \multirow{2}{*}{Model} & \multicolumn{4}{c}{MiniLM} & \multicolumn{4}{c}{MPNet} & \multicolumn{4}{c}{BERTScore} \\
    \cmidrule(lr){3-6} \cmidrule(lr){7-10} \cmidrule(lr){11-14}
                          & & ES   & FR   & KO   & JA   & ES   & FR   & KO   & JA   & ES   & FR   & KO   & JA \\
    \midrule
    \multicolumn{2}{c}{} & \multicolumn{12}{c}{\textbf{English Lyrics $\longleftrightarrow$ Translated Lyrics}} \\
    \midrule
    & Human Expert    & 0.1804 & 0.1622 & 0.0755 & 0.0801 & 0.6392 & 0.6126 & 0.5746 & 0.5374 & 0.6856 & 0.6867 & 0.6486 & 0.6383 \\[0.5ex]
    \cdashline{2-14}
    \noalign{\vskip 0.5ex}
    \multirow{5}{*}{\rotatebox{90}{w/o SC}} 
    & Google Translate    & \textbf{0.2290} & 0.2322 & \textbf{0.0912} & 0.0810 & \textbf{0.9052} & \textbf{0.9027} & \textbf{0.8572} & \textbf{0.8569} & \textbf{0.7751} & \textbf{0.7811} & 0.6805 & 0.6743 \\
    & MBart-50            & 0.1993 & \textbf{0.2491} & 0.0856 & \textbf{0.0994} & 0.7658 & 0.8895 & 0.8496 & 0.8415 & 0.7331 & 0.7775 & 0.6677 & \textbf{0.6828} \\
    & Qwen-72B            & 0.2174 & 0.2258 & 0.0657 & 0.0726 & 0.9004 & 0.8987 & 0.8412 & 0.8362 & 0.7725 & 0.7771 & \textbf{0.6849} & 0.6784 \\
    & GPT-4o              & 0.2131 & 0.2167 & 0.0628 & 0.0706 & 0.8993 & 0.8969 & 0.8486 & 0.8422 & 0.7688 & 0.7727 & 0.6812 & 0.6747 \\
    & Gemini 2.0 & 0.2061 & 0.2146 & 0.0687 & 0.0725 & 0.8931 & 0.8911 & 0.8426 & 0.8337 & 0.7575 & 0.7624 & 0.6747 & 0.6621 \\
    \cdashline{2-14}
    \noalign{\vskip 0.5ex}
    \multirow{4}{*}{\rotatebox{90}{w/ SC}} 
    & Qwen-72B & 0.2042 & 0.2046 & 0.0653 & 0.0778 & 0.8825 & 0.8776 & 0.8226 & 0.8103 & 0.7615 & 0.7656 & 0.6812 & 0.6764 \\
    & GPT-4o & 0.2042 & 0.2047 & 0.0703 & 0.0837 & 0.8295 & 0.8417 & 0.7932 & 0.7818 & 0.7421 & 0.7542 & 0.6759 & 0.6711 \\
    & Gemini 2.0 & 0.1857 & 0.1785 & 0.0673 & 0.0826 & 0.7813 & 0.7904 & 0.7544 & 0.7327 & 0.7169 & 0.7234 & 0.6686 & 0.6604 \\
    & SylAVL-CoT (Gemini 2.0)         & 0.1900 & 0.1849 & 0.0679 & 0.0867 & 0.7652 & 0.7859 & 0.7289 & 0.6953 & 0.7142 & 0.7278 & 0.6635 & 0.6567 \\
    \midrule
    \multicolumn{2}{c}{} & \multicolumn{12}{c}{\textbf{Dubbed Lyrics $\longleftrightarrow$ Translated Lyrics}} \\
    \midrule
    \multirow{5}{*}{\rotatebox{90}{w/o SC}} & Google Translate    & 0.5584 & 0.5092 & 0.6661 & 0.4619 & 0.6724 & 0.6490 & 0.6177 & 0.5995 & 0.7402 & 0.7348 & 0.7158 & 0.6824 \\
    & MBart-50            & 0.4938 & 0.4891 & 0.6671 & 0.4547 & 0.5764 & 0.6339 & 0.6075 & 0.6123 & 0.7031 & 0.7258 & 0.7036 & 0.6925 \\
    & Qwen-72B            & 0.5622 & 0.5145 & 0.6920 & 0.4820 & \textbf{0.6772} & \textbf{0.6549} & 0.6371 & 0.6329 & 0.7423 & \textbf{0.7384} & 0.7359 & 0.7012 \\
    & GPT-4o              & 0.5564 & 0.5001 & 0.6932 & 0.4872 & 0.6703 & 0.6354 & 0.6332 & 0.6244 & 0.7386 & 0.7321 & 0.7353 & 0.7005 \\
    & Gemini 2.0 & \textbf{0.5623} & \textbf{0.5200} & 0.6946 & 0.4794 & 0.6709 & 0.6516 & 0.6359 & 0.6104 & 0.7328 & 0.7332 & 0.7318 & 0.6885 \\
    \cdashline{2-14}
    \noalign{\vskip 0.5ex}
    \multirow{4}{*}{\rotatebox{90}{w/ SC}} 
    & Qwen-72B & 0.5619 & 0.4934 & 0.6993 & 0.4905 & 0.6729 & 0.6323 & 0.6448 & 0.6489 & 0.7410 & 0.7309 & 0.7416 & 0.7118 \\
    & GPT-4o & 0.5573 & 0.5107 & \textbf{0.7085} & \textbf{0.5042} & 0.6722 & 0.6536 & 0.6472 & 0.6715 & \textbf{0.7450} & 0.7380 & 0.7530 & 0.7238 \\
    & Gemini 2.0 & 0.5383 & 0.4910 & 0.6979 & 0.4871 & 0.6538 & 0.6341 & \textbf{0.6506} & 0.6687 & 0.7314 & 0.7224 & 0.7468 & 0.7152 \\
    & SylAVL-CoT (Gemini 2.0)         & 0.5395 & 0.4975 & 0.7016 & 0.4975 & 0.6561 & 0.6402 & 0.6467 & \textbf{0.6814} & 0.7358 & 0.7300 & \textbf{0.7509} & \textbf{0.7241} \\
    \bottomrule
    \end{tabular}%
    }
    \caption{\textbf{Comparison with other models: Sense.} \textbf{Best} are highlighted.}
    \label{tab:suppl_sem_metric}
    \end{table*}
\begin{table*}[htbp]
\centering
\scriptsize               
\setlength{\tabcolsep}{4pt} 
\resizebox{\textwidth}{!}{%
\begin{tabular}{c *{4}{c} | *{4}{c} | *{4}{c}}
\toprule
\multirow{2}{*}{Modality} & \multicolumn{4}{c}{MiniLM} & \multicolumn{4}{c}{MPNet} & \multicolumn{4}{c}{BERTScore} \\
\cmidrule(lr){2-5} \cmidrule(lr){6-9} \cmidrule(lr){10-13}
                      & ES   & FR   & KO   & JA   & ES   & FR   & KO   & JA   & ES   & FR   & KO   & JA \\
\midrule
\multicolumn{1}{c}{} & \multicolumn{12}{c}{\textbf{English Lyrics $\longleftrightarrow$ Translated Lyrics}} \\ \midrule
T & 0.1872 & 0.1846 & \textbf{0.0692} & \underline{0.0894} & 0.7584 & \underline{0.7888} & \underline{0.7391} & \underline{0.6965} & 0.7082 & 0.7265 & \underline{0.6651} & \underline{0.6592} \\
T + V  & 0.1871 & 0.1788 & \underline{0.0692} & 0.0849 & 0.7586 & 0.7794 & 0.7286 & 0.6952 & 0.7098 & 0.7236 & 0.6630 & 0.6572 \\
T + A   & \textbf{0.1921} & \textbf{0.1874} & 0.0688 & \textbf{0.0897} & \textbf{0.7723} & \textbf{0.8020} & \textbf{0.7484} & \textbf{0.7133} & \textbf{0.7143} & \textbf{0.7320} & \textbf{0.6661} & \textbf{0.6608} \\
T + A + V & \underline{0.1900} & \underline{0.1849} & 0.0679 & 0.0867 & \underline{0.7652} & 0.7859 & 0.7298 & 0.6953 & \underline{0.7142} & \underline{0.7278} & 0.6635 & 0.6567 \\
\midrule
\multicolumn{1}{c}{} & \multicolumn{12}{c}{\textbf{Dubbed Lyrics $\longleftrightarrow$ Translated Lyrics}} \\ \midrule
T & 0.5252 & 0.4869 & 0.6969 & 0.4874 & 0.6481 & 0.6332 & 0.6398 & 0.6748 & 0.7296 & 0.7245 & 0.7480 & 0.7243 \\
T + V     & 0.5346 & 0.4878 & 0.6995 & \textbf{0.4929} & 0.6483 & 0.6284 & 0.6433 & \underline{0.6796} & 0.7315 & 0.7231 & 0.7489 & 0.7256 \\
T + A     & \underline{0.5382} & \underline{0.4918} & \underline{0.6995} & \underline{0.4895} & \underline{0.6559} & \underline{0.6400} & \underline{0.6462} & 0.6785 & \underline{0.7345} & \underline{0.7278} & \underline{0.7500} & \textbf{0.7262} \\
T + A + V    & \textbf{0.5395} & \textbf{0.4975} & \textbf{0.7016} & 0.4893 & \textbf{0.6561} & \textbf{0.6402} & \textbf{0.6466} & \textbf{0.6814} & \textbf{0.7349} & \textbf{0.7285} & \textbf{0.7509} & \underline{0.7260} \\
\bottomrule
\end{tabular}%
}
\caption{\textbf{Combined metrics of Ablation studies for MiniLM, MPNet, and BERTScore.}}
\label{tab:abl_combined_metrics}
\end{table*}

\begin{figure*}
    \centering
    \includegraphics[width=\linewidth]{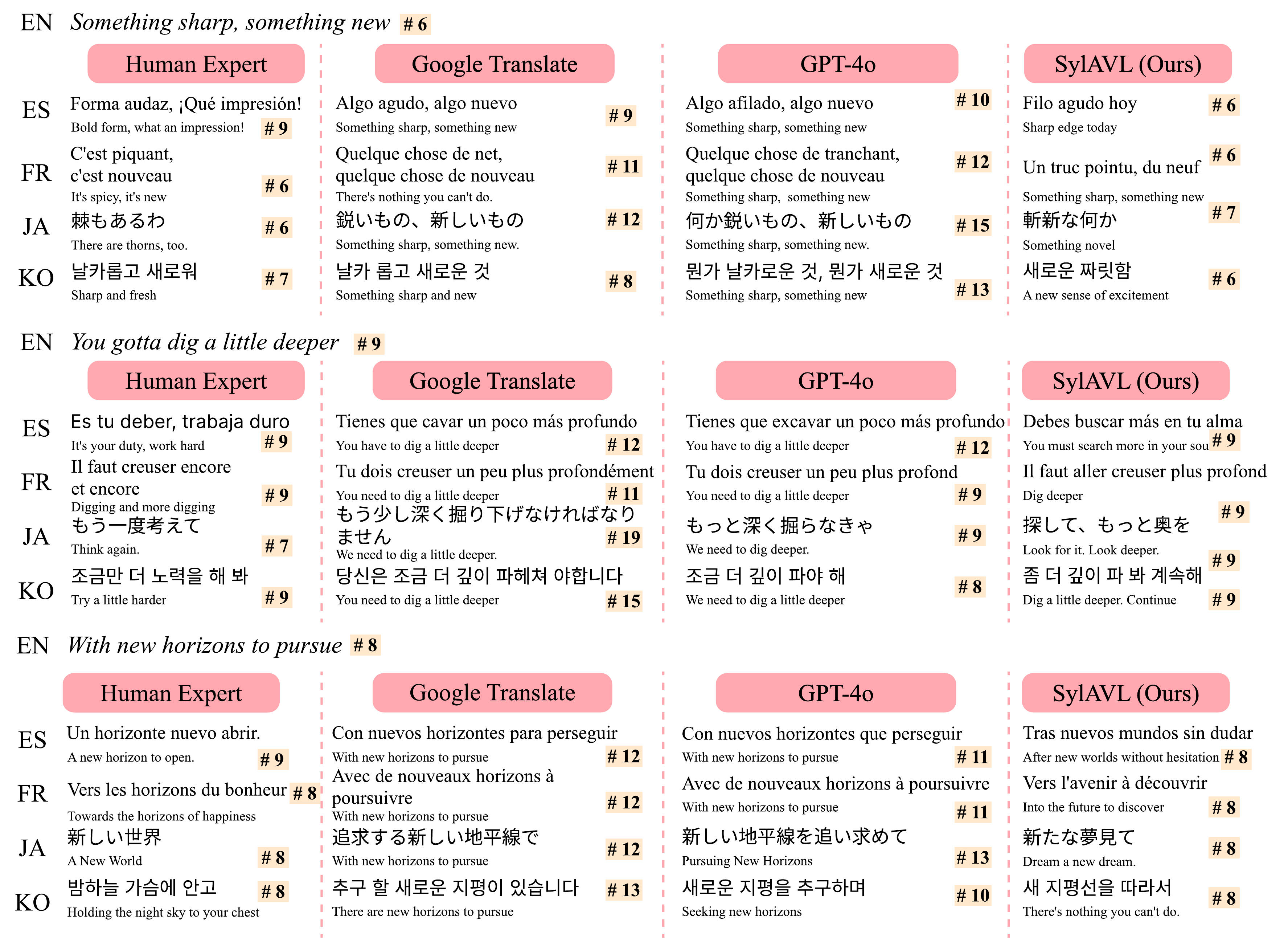}
    \caption{\textbf{Qualitative results}. This figure showcases translations of English lyrics into Spanish (ES), French (FR), Japanese (JA), and Korean (KO) by Human Experts, Google Translate, GPT-4o, and \textbf{\texttt{SylAVL-CoT}} (Ours). Notably, the translations from \textbf{\texttt{SylAVL-CoT}} (Ours) demonstrate syllable counts (where \# denotes the syllable counts) that are most similar to those of the original English lyrics. The examples also allow for a comparison of how specific English lyric lines are rendered by Human Experts versus our \textbf{\texttt{SylAVL-CoT}} model.}
    \label{fig:appendix_qual}
\end{figure*}

\section{Additional Experiments Details}
\label{appendix:additional_exp}
\subsection{User Study Details}
\label{appendix:userstudy-detail}
\begin{figure*}[htb!]
    \centering
    \includegraphics[width=1.0\linewidth]{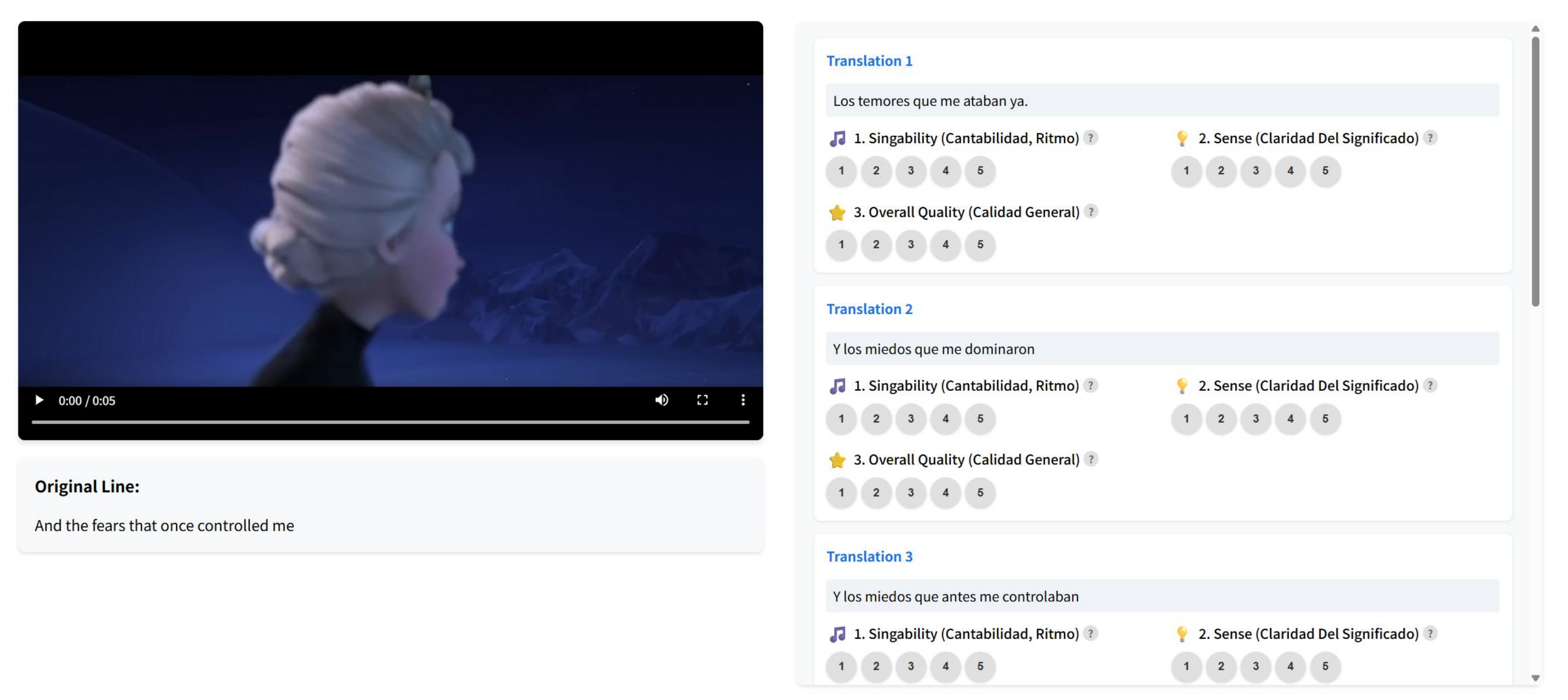}
    \caption{\textbf{Example of User Study Website}}
    \label{fig:userstudy-example}
\end{figure*}
For user study, we developed a dedicated website where participants could watch the 10 original English video clips alongside the translated lyrics from different systems. To ensure a clear understanding of the evaluation task and to guide participants towards evaluations aligned with our intended criteria, we provided comprehensive instructions before they began. These instructions included a detailed outline of the tasks to be performed, clear definitions and criteria for each evaluation metric, and examples. This preparation aimed to equip participants to make informed and consistent judgments.

Specifically, before commencing the evaluation, participants were presented with an "Evaluation Guide." This guide, titled "Evaluation Guide," first outlined its purpose: "On this page, you will evaluate the quality of song lyrics translation." It then provided the following key instructions:
\begin{itemize}
    \item Compare the original lyrics with the translated lyrics.
    \item Evaluate the translated lyrics along with the melody of the song through the provided video.
\end{itemize}
Participants were asked to evaluate the translations based on three criteria:
\begin{itemize}
    \item \textbf{Singability:} Evaluate how well the translated lyrics fit with the melody. If possible, try singing them yourself.
    \item \textbf{Sense:} Evaluate how clearly and accurately the translated lyrics convey the original meaning.
    \item \textbf{Overall Quality:} Evaluate the overall quality of the translation.
\end{itemize}
The scoring mechanism was explained as: "Rate each translation from 1 to 5. (1: Very poor, 2: Poor, 3: Average, 4: Good, 5: Very good)". This entire guide was translated and presented to participants in their respective evaluation languages. Finally, participants would click a "Start Evaluation" button to proceed.
Example image of user study website can be seen in Figure~\ref{fig:userstudy-example}.

Each evaluation session lasted approximately 30 minutes and involved 10 participants per language. Participants were compensated at a rate of approximately €5 each for their time and effort.

\subsection{Qualitative Results}
\label{appendix:qual_results}
We present various qualitative results, as shown in Figure~\ref{fig:appendix_qual}. \textbf{\texttt{SylAVL-CoT}} preserves both the original meaning and syllable structure, ensuring singability.

\subsection{Semantic Evaluation}
\noindent \textbf{Comparison with other models.} We conducted comparative experiments using the metrics employed in previous studies to measure semantic fidelity. This is presented in Table~\ref{tab:suppl_sem_metric}. For the translation between English and non-English, Google Translate scored the highest score with the overall languages. As human expert considers various factors for translating the lyrics, the performance is the lowest among various semantic evaluation metrics. When comparing dubbed and translated lyrics, our proposed model achieved high performance for a wide range of languages. On the other hand, Google Translate's translation performance seems to be relatively poor.

\noindent \textbf{Ablation study.} 
To demonstrate the usefulness of multimodality, we conducted a semantic evaluation based on different modalities, as shown in Table~\ref{tab:abl_combined_metrics}. As observed in Table~\ref{tab:abl_combined_metrics}, incorporating multimodal information alongside text leads to better performance.

\subsection{Cross-lingual \textbf{\texttt{SylAVL-CoT}} Experiments}
\label{appendix:cross-lingual}
The MAVL dataset is inherently well-suited for comprehensive cross-lingual benchmarking. Most of its data entries across all supported languages (English, Spanish, French, Korean, and Japanese) include aligned video and meticulously synchronized lyrics. This rich, multimodal, and parallel structure enables the evaluation of translation not only from a common source language (like English) to various target languages but also between any pair of the supported languages, or from non-English languages back to English. This flexibility allows for a deeper understanding of a model's translation capabilities across diverse linguistic landscapes.

To illustrate this, we conducted additional cross-lingual experiments with \textbf{\texttt{SylAVL-CoT}}, evaluating its performance when translating between different language pairs beyond the English-centric evaluations presented in the main paper. The Syllable Error (SE), Syllable Distance and MPNet semantic similarity scores for these experiments are presented in Table~\ref{tab:app_AL_SE} and Table~\ref{tab:app_AL_mpnet}, respectively.

\begin{table*}[!t]
    \centering
    \scriptsize
    \setlength{\tabcolsep}{2pt}
    \renewcommand{\arraystretch}{1.2}
    \resizebox{\textwidth}{!}{%
    \begin{tabular}{c *{16}{>{\centering\arraybackslash}p{0.7cm}} }
    \toprule
    Metric & \multicolumn{16}{c}{Syllable Error $\downarrow$} \\
    \midrule
     Source & \multicolumn{4}{c}{ES} & \multicolumn{4}{c}{FR} & \multicolumn{4}{c}{KO} & \multicolumn{4}{c}{JA} \\
    \cmidrule(lr){2-5}\cmidrule(lr){6-9}\cmidrule(lr){10-13}\cmidrule(lr){14-17}
    Target &  EN & FR & KO & JA & EN & ES & KO & JA & EN & ES & FR & KO & EN & ES & FR & JA\\
    \midrule
    \multicolumn{1}{c}{} & \multicolumn{16}{c}{{\textbf{English Lyrics $\longleftrightarrow$ Translated Lyrics}}} \\
    \midrule
    Human Expert          & 1.005 & 1.280 & 1.221 & 1.740 & 0.824 & 1.945 & 1.225 & 1.772 & 0.659 & 1.743 & 1.099 & 1.475 & 1.321 & 2.239 & 1.617 & 1.454 \\[0.5ex]
    \cdashline{1-17}
    \noalign{\vskip 0.5ex}
    Google Translate      & 2.154 & 1.994 & 7.021 & 13.106 & 1.791 & 5.042 & 9.306 & 16.869 & 2.631 & 3.114 & 2.530 & 8.991 & 3.155 & 2.821 & 2.901 & 3.833 \\[0.5ex]
    \cdashline{1-17}
     Gemini 2.0 (w/ SC)               & 1.442 & 1.404 & 1.169 & 3.443 & 0.773 & 3.105 & 2.021 & 5.229 & 1.301 & 1.634 & 1.274 & 4.023 & 2.152 & 1.191 & 1.983 & 1.280 \\[0.5ex]
     SylAVL-CoT (Gemini 2.0)        & \textbf{0.298} & \textbf{0.965} & \textbf{0.382} & \textbf{1.640} & \textbf{0.353} & \textbf{1.067} & \textbf{0.613} & \textbf{2.050} & \textbf{0.311} & \textbf{0.795} & \textbf{0.932} & \textbf{1.704} & \textbf{0.333} & \textbf{0.646} & \textbf{1.053} & \textbf{0.348} \\[0.5ex]
    \midrule
    \multicolumn{1}{c}{} & \multicolumn{16}{c}{\textbf{Dubbed Lyrics $\longleftrightarrow$ Translated Lyrics}} \\
    \midrule
     Google Translate      & 1.838 & 2.265 & 7.886 & 14.200 & 1.770 & 4.267 & 9.157 & 16.662 & 2.554 & 3.246 & 2.732 & 9.213 & 2.996 & 3.096 & 3.001 & 4.017 \\[0.5ex]
    \cdashline{1-17}
    Gemini 2.0 (w/ SC)        & \textbf{1.030} & \textbf{1.412} & \textbf{1.854} & 4.718 & \textbf{0.997} & 2.519 & 2.280 & 5.339 & 1.397 & 1.896 & 1.534 & 4.364 & 2.107 & 2.033 & 2.137 & 1.631 \\[0.5ex]
    SylAVL-CoT (Gemini 2.0)         & 1.780 & 1.746 & 1.966 & \textbf{3.173} & 1.062 & \textbf{1.658} & \textbf{1.444} & \textbf{2.848} & \textbf{1.035} & \textbf{1.504} & \textbf{1.501} & \textbf{2.405} & \textbf{1.527} & \textbf{1.810} & \textbf{1.905} & \textbf{1.514} \\[0.5ex]
    \bottomrule
    \end{tabular}%
    }
    \caption{\textbf{Comparison of Syllable Error (SE) in Cross-Lingual Lyrics Translation for various Source Languages (ES, FR, KO, JA).}}
    \label{tab:app_AL_SE}
    \end{table*}
    
\begin{table*}[!t]
    \centering
    \scriptsize
    \setlength{\tabcolsep}{2pt}
    \renewcommand{\arraystretch}{1.2}
    \resizebox{\textwidth}{!}{%
    \begin{tabular}{c *{16}{>{\centering\arraybackslash}p{0.7cm}} }
    \toprule
    Metric & \multicolumn{16}{c}{Syllable Distance $\downarrow$} \\
    \midrule
     Source & \multicolumn{4}{c}{ES} & \multicolumn{4}{c}{FR} & \multicolumn{4}{c}{KO} & \multicolumn{4}{c}{JA} \\
    \cmidrule(lr){2-5}\cmidrule(lr){6-9}\cmidrule(lr){10-13}\cmidrule(lr){14-17}
    Target &  EN & FR & KO & JA & EN & ES & KO & JA & EN & ES & FR & KO & EN & ES & FR & JA\\
    \midrule
    \multicolumn{1}{c}{} & \multicolumn{16}{c}{{\textbf{English Lyrics $\longleftrightarrow$ Translated Lyrics}}} \\
    \midrule
    Human Expert          & 0.116 & 0.136 & 0.132 & 0.172 & 0.078 & 0.136 & 0.153 & 0.108 & 0.082 & 0.132 & 0.108 & 0.138 & 0.133 & 0.172 & 0.153 & 0.138 \\[0.5ex]
    \cdashline{1-17}
    \noalign{\vskip 0.5ex}
    Google Translate      & 0.303 & 0.237 & 0.609 & 0.398 & 0.204 & 0.300 & 0.779 & 0.497 & 0.474 & 0.309 & 0.387 & 0.484 & 0.557 & 0.325 & 0.438 & 0.336 \\[0.5ex]
    \cdashline{1-17}
     Gemini 2.0 (w/ SC)               & 0.174 & 0.168 & 0.202 & 0.102 & 0.077 & 0.182 & 0.300 & 0.133 & 0.156 & 0.116 & 0.150 & 0.235 & 0.295 & 0.135 & 0.277 & 0.148 \\[0.5ex]
     SylAVL-CoT (Gemini 2.0)        & \textbf{0.028} & \textbf{0.104} & \textbf{0.109} & \textbf{0.029} & \textbf{0.027} & \textbf{0.072} & \textbf{0.138} & \textbf{0.045} & \textbf{0.030} & \textbf{0.058} & \textbf{0.105} & \textbf{0.129} & \textbf{0.041} & \textbf{0.053} & \textbf{0.130} & \textbf{0.040} \\[0.5ex]
    \midrule
    \multicolumn{1}{c}{} & \multicolumn{16}{c}{\textbf{Dubbed Lyrics $\longleftrightarrow$ Translated Lyrics}} \\
    \midrule
     Google Translate      & 0.251 & 0.236 & 0.699 & 0.461 & 0.201 & 0.267 & 0.773 & 0.492 & 0.455 & 0.352 & 0.402 & 0.518 & 0.531 & 0.376 & 0.445 & 0.361 \\[0.5ex]
    \cdashline{1-17}
    Gemini 2.0 (w/ SC)                & 0.126 & 0.156 & 0.290 & 0.154 & 0.095 & 0.165 & 0.319 & 0.166 & 0.167 & 0.160 & 0.175 & 0.271 & 0.296 & 0.231 & 0.296 & 0.191 \\[0.5ex]
    SylAVL-CoT (Gemini 2.0)        & \textbf{0.120} & \textbf{0.153} & \textbf{0.221} & \textbf{0.144} & \textbf{0.086} & \textbf{0.146} & \textbf{0.216} & \textbf{0.128} & \textbf{0.146} & \textbf{0.192} & \textbf{0.165} & \textbf{0.266} & \textbf{0.138} & \textbf{0.177} & \textbf{0.207} & \textbf{0.150} \\[0.5ex]
    \bottomrule
    \end{tabular}%
    }
    \caption{\textbf{Comparison of Syllable Distance (SD) in Cross-Lingual Lyrics Translation for various Source Languages (ES, FR, KO, JA).}}
    \label{tab:app_AL_SD}
    \end{table*}
    
\begin{table*}[!t]
    \centering
    \scriptsize
    \setlength{\tabcolsep}{2pt}
    \renewcommand{\arraystretch}{1.2}
    \resizebox{\textwidth}{!}{%
    \begin{tabular}{c *{16}{>{\centering\arraybackslash}p{0.7cm}} }
    \toprule
    Metric & \multicolumn{16}{c}{MPNet Score $\uparrow$} \\
    \midrule
     Source & \multicolumn{4}{c}{ES} & \multicolumn{4}{c}{FR} & \multicolumn{4}{c}{KO} & \multicolumn{4}{c}{JA} \\
    \cmidrule(lr){2-5}\cmidrule(lr){6-9}\cmidrule(lr){10-13}\cmidrule(lr){14-17}
    Target &  EN & FR & KO & JA & EN & ES & KO & JA & EN & ES & FR & KO & EN & ES & FR & JA\\
    \midrule
    \multicolumn{1}{c}{} & \multicolumn{16}{c}{{\textbf{English Lyrics $\longleftrightarrow$ Translated Lyrics}}} \\
    \midrule
    Human Expert          & 0.639 & 0.578 & 0.580 & 0.574 & 0.609 & 0.578 & 0.561 & 0.565 & 0.582 & 0.580 & 0.561 & 0.617 & 0.535 & 0.574 & 0.565 & 0.617 \\[0.5ex]
    \cdashline{1-17}
    \noalign{\vskip 0.5ex}
    Google Translate      & \textbf{0.884} & \textbf{0.909} & \textbf{0.857} & \textbf{0.863} & \textbf{0.897} & \textbf{0.920} & \textbf{0.871} & \textbf{0.879} & \textbf{0.812} & \textbf{0.837} & \textbf{0.839} & \textbf{0.865} & \textbf{0.781} & \textbf{0.824} & \textbf{0.831} & 0.848 \\[0.5ex]
    \cdashline{1-17}
     Gemini 2.0 (w/ SC)                & 0.809 & 0.833 & 0.800 & 0.798 & 0.823 & 0.850 & 0.803 & 0.803 & 0.767 & 0.799 & 0.790 & 0.862 & 0.736 & 0.792 & 0.785 & \textbf{0.857} \\[0.5ex]
     SylAVL-CoT (Gemini 2.0)         & 0.762 & 0.804 & 0.783 & 0.757 & 0.776 & 0.810 & 0.779 & 0.759 & 0.713 & 0.768 & 0.766 & 0.804 & 0.691 & 0.759 & 0.761 & 0.837 \\[0.5ex]
    \midrule
    \multicolumn{1}{c}{} & \multicolumn{16}{c}{\textbf{Dubbed Lyrics $\longleftrightarrow$ Translated Lyrics}} \\
    \midrule
     Google Translate      & 0.590 & 0.561 & 0.556 & 0.565 & 0.563 & 0.552 & 0.536 & 0.550 & 0.522 & 0.527 & 0.517 & 0.590 & 0.460 & 0.498 & 0.494 & 0.536 \\[0.5ex]
    \cdashline{1-17}
    Gemini 2.0 (w/ SC)                & 0.597 & 0.570 & \textbf{0.600} & 0.641 & \textbf{0.563} & 0.551 & 0.568 & 0.620 & \textbf{0.545} & \textbf{0.546} & 0.530 & 0.670 & 0.464 & 0.504 & 0.494 & 0.574 \\[0.5ex]
    SylAVL-CoT (Gemini 2.0)        & \textbf{0.599} & \textbf{0.576} & 0.598 & \textbf{0.662} & 0.560 & \textbf{0.552} & \textbf{0.574} & \textbf{0.644} & 0.541 & 0.544 & \textbf{0.541} & \textbf{0.676} & \textbf{0.473} & \textbf{0.505} & \textbf{0.508} & \textbf{0.567} \\[0.5ex]
    \bottomrule
    \end{tabular}%
    }
    \caption{\textbf{Comparison of MPNet Score in Cross-Lingual Lyrics Translation for various Source Languages (ES, FR, KO, JA).}}
    \label{tab:app_AL_mpnet}
    \end{table*}

Upon analyzing the results, it is observed that \textbf{\texttt{SylAVL-CoT}} generally demonstrates a performance profile consistent with that reported for English-to-other-language translations in the main body of this paper (refer to Table~\ref{tab:comp_syllable} for syllable error metrics and Table~\ref{tab:base_metrics_mpnet_vertical} for semantic scores). While minor variations naturally occur depending on the specific linguistic characteristics and distance between language pairs, the overall efficacy of \textbf{\texttt{SylAVL-CoT}} in maintaining both singability and semantic coherence remains evident across these broader cross-lingual scenarios. This underscores the robustness of the \textbf{\texttt{SylAVL-CoT}} approach and the utility of the MAVL dataset for multifaceted translation evaluation.
Notably, when Spanish is the source language, a relatively higher Syllable Error (SE) with Dubbed Lyrics can be observed. This aligns with the observation from Table~\ref{tab:comp_syllable}, where Human Expert translations into Spanish also showed a high SE. This suggests that Spanish lyrics tend to have a higher syllable count per line compared to the original English lyrics. Consequently, translations generated by \textbf{\texttt{SylAVL-CoT}} from a Spanish source, which aim for low Syllable Error and Syllable Distance against the Spanish source, might naturally reflect this higher syllable count distribution. If the Dubbed Lyrics (against which the comparison is made) have a syllable count closer to the original English (i.e., generally lower), this would explain the increased SE when comparing \textbf{\texttt{SylAVL-CoT}}'s Spanish-sourced output to these Dubbed Lyrics.

\section{Details of MAVL Dataset}
\label{appendix:appendix_dataset}
\subsection{Details of Lyrics Collection by Web Crawling}
\begin{figure*}
    \centering
    \includegraphics[width=\linewidth]{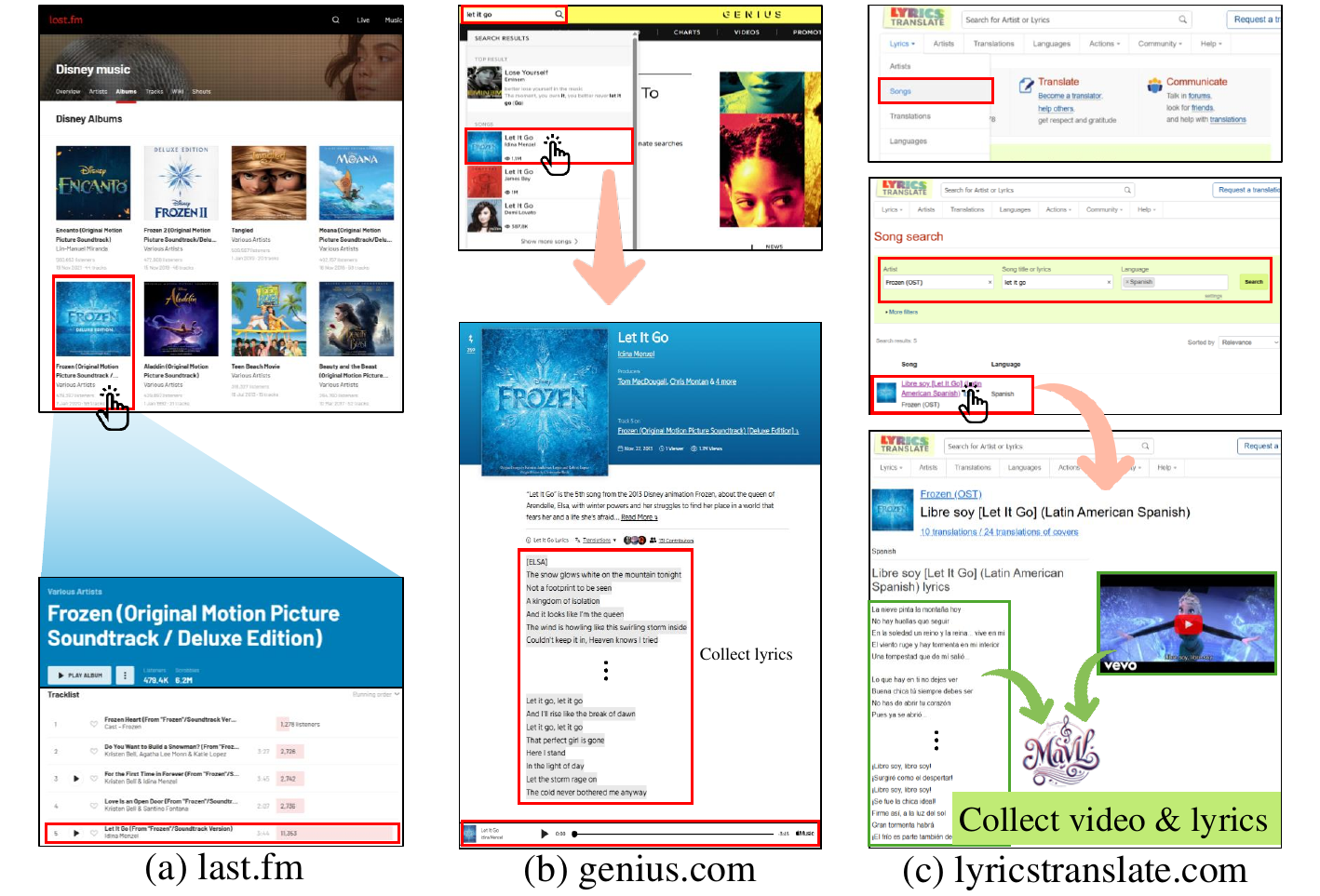}
    \caption{\textbf{Web crawling process for MAVL dataset collection.} Our collection pipeline proceeds in three steps—(a) $\rightarrow$ (b) $\rightarrow$ (c)—each corresponding to a specific website used in the process. This example is from the Spanish version of the OST "Let It Go" from Frozen, produced by Disney.}
    \label{fig:crawling_mavl_dataset}
\end{figure*}
This section describes how we collected multilingual lyrics and corresponding videos, as illustrated in Figure~\ref{fig:crawling_mavl_dataset}.
We began by gathering metadata for animated songs from \texttt{last.fm} (Figure~\ref{fig:crawling_mavl_dataset}-(a)), followed by collecting English lyrics from \texttt{genius} (Figure~\ref{fig:crawling_mavl_dataset}-(b)).
Based on this information, we searched \texttt{lyricstranslate.com} by country to collect localized lyrics and corresponding videos in multiple languages (Figure~\ref{fig:crawling_mavl_dataset}-(c)).

\subsection{Alignment Methodology}
\label{appendix:alignment_methodology}
To maximize alignment accuracy between lyrics, audio, and video, we employed a comprehensive approach using \texttt{stable-ts}, which builds upon the Whisper model~\cite{radford2022robustspeechrecognitionlargescale}. We chose stable-ts over alternatives like WhisperX because it is specifically designed for aligning existing text with audio, rather than just creating timestamps, and it provides robust multilingual support critical for our diverse language set.

Our alignment process incorporates several quality assurance techniques:
\begin{itemize}
    \item \textbf{Ensemble Approach}: We used an ensemble of Whisper-large-v1 and v2 models (v3 was found to be less reliable for this task), selecting the best alignment based on confidence scores.
    \item \textbf{Vocal Separation}: We employed the DEMUCS model to separate vocals from background music, performing alignment on both the original track and the vocal-only track. If the original track alignment failed, we used the vocal-only track alignment as a fallback.
    \item \textbf{Quality Filtering}: Alignments with low confidence scores or significant timing discrepancies were excluded from the dataset to ensure high quality.
\end{itemize}

\subsection{MAVL Dataset Format}
\begin{figure*}
    \centering
    \includegraphics[width=1.0\linewidth]{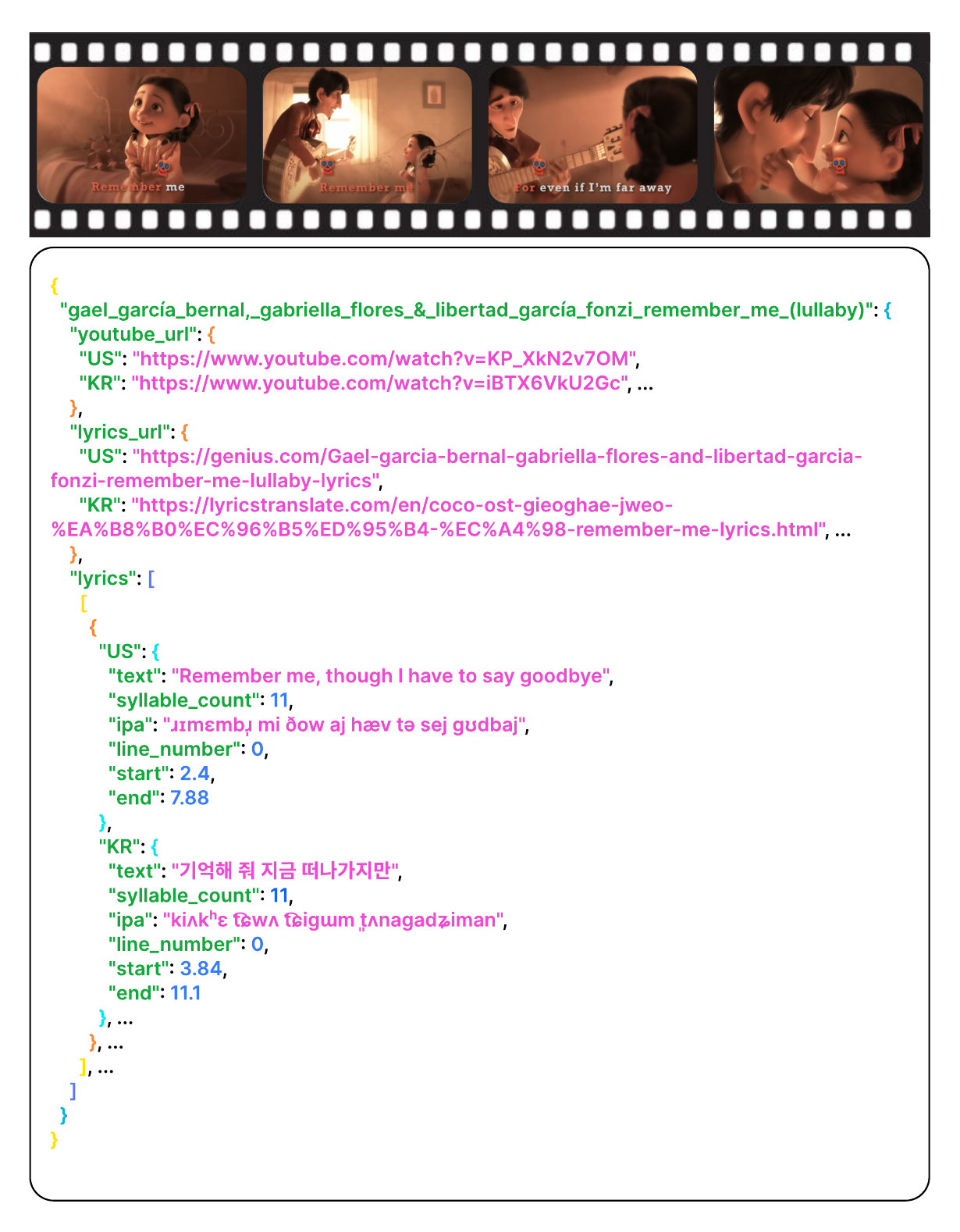}
    \caption{\textbf{MAVL Dataset format.} This is an example of an annotation for a single song. We provide the MAVL dataset in JSON format. This example is ``Remember me'' from ``COCO'' by Disney.}
    \label{fig:dataset_format}
\end{figure*}

\begin{CJK}{UTF8}{gbsn}
In order to comply with copyright regulations, we only provide URLs rather than distributing the full lyrics, and we reconstruct each line for alignment by extracting a compact representation. For instance, for the English line “Remember me though I have to say goodbye,” we split it by spaces and record the first letter of each word (R, m, t, I, h, t, s, g) along with the line’s first and last words (“Remember” and “goodbye”), resulting in a representation like [“RmtIhtsg”, “Remember”, “goodbye”]. With Japanese lyrics—where spacing does not naturally separate words—we use morphological analysis (e.g., MeCab\footnote{https://github.com/SamuraiT/mecab-python3}) to split the line into tokens. We then combine these tokens in pairs, which we treat similarly to the English case by storing partial data (such as the concatenation of first letters or selected tokens) to enable accurate restoration of the original line once the corresponding URL is accessed.
\end{CJK}

Upon finalizing line reconstruction, IPA transcription, and syllable counting, the dataset assumes the structure depicted in Figure~\ref{fig:dataset_format}.

\section{Why we choose CoT method}
Building an end-to-end multimodal lyrics translation system that handles text, audio, and video poses several unique challenges. Multilingual lyrics datasets with aligned audiovisual content are extremely rare. Training a dedicated model from scratch on such limited data is a significant hurdle. Even if such a model were developed, the training process itself would be time-consuming and resource-intensive. Furthermore, adapting the model to new languages would require substantial effort in curating new aligned datasets and retraining, making the system inflexible to evolving linguistic needs.

To address these constraints, we leverage \textbf{Gemini 2.0 Flash}~\cite{gemini20flash}, a closed-source MLLM capable of processing audio and video inputs for all six languages. By employing Gemini 2.0 Flash, we bypass the need to develop and train a new model on a highly specialized, low-resource task. Instead, we explore whether prompt-based techniques alone can effectively solve the lyrics translation problem, even for content that requires multimodal understanding.

Our approach centers on adapting Gemini 2.0 Flash via \emph{prompt tuning} rather than extensive fine-tuning. Specifically, we propose a \emph{Multimodal Chain-of-Thought} pipeline that augments standard chain-of-thought reasoning with additional cues derived from audio and video data. This design allows the model to incorporate contextual information from multiple modalities, which is crucial for translation tasks involving music, animation clips, and other audiovisual elements.

\section{Prompts for Lyrics Translation}
\label{appendix:cot_prompts}
We provide the prompt used by the \textbf{\texttt{SylAVL-CoT}} model in Table~\ref{tab:prompt_CoT_song_translation}.
The prompt for syllable-constrained lyrics translation is in Table~\ref{tab:prompt_SC_song_translation}.
The prompt for text-only \textbf{\texttt{SylAVL-CoT}} prompt used in ablation Table~\ref{tab:ablation-mpnet} is in Table~\ref{tab:prompt_CoT_text_only}. We have not included the specific prompts for audio-only and video-only \textbf{\texttt{SylAVL-CoT}} ablations, as these are derived directly by combining elements from the text-only and the complete audiovisual \textbf{\texttt{SylAVL-CoT}} prompts.
\begin{CJK}{UTF8}{mj}
\begin{table*}[ht!]
\centering
\fontsize{9pt}{9.5pt}\selectfont
\noindent\fbox{
\begin{minipage}{\dimexpr\linewidth-2\fboxsep-2\fboxrule}
\tt
You are a professional song translator with expertise in preserving musicality. Translate the following song lyrics from \{source\_lang\} to \{target\_lang\}.\\
Please perform the following steps:\\
{\color{blue}
1.\quad Identify the Core Lyric and Perform Syllable Segmentation\\
\quad - You are given both an audio clip (which may include additional lyrics before or after) and the corresponding lyric text for a specific scene.\\
\quad - Use the provided lyric text to determine the exact lyric line you need to process, and disregard any extraneous audio content that is not part of the given text.\\
\quad - Listen carefully to the provided audio to capture the natural rhythm, pronunciation, and any important phonetic, musical, or syllabic characteristics of the identified lyric line. If such notable features exist, provide a brief explanation of them before proceeding with the translation.\\
\quad - You will be also given the real syllable count of the original lyric.\\
\quad - Break down the determined lyric line into its constituent syllables based on the audio's natural breaks and real syllable count.\\
\quad - Ensure that each syllable is logically segmented according to the pronunciation. And write it down in the following format:\\
\quad\quad Example:\\
\quad\quad - If the lyric is: "Three months of winter coolness and awesome holidays" \\
\quad\quad A correct segmentation might be:\\
\quad\quad \hspace*{1em} ["Three", "months", "of", "win", "ter", "cool", "ness"]\\
\quad\quad This original syllable list has 7 syllables.\\
}

{\color{magenta}
2.\quad Generate the Target Language Translation Syllable List Utilizing the Video Information\\
\quad - Translate the meaning of the lyric naturally and idiomatically into the target language (\{target\_lang\}).\\
\quad - Review the provided video context and generate the description of the video to understand the intended mood, imagery, and cultural nuances of the original lyric.\\
\quad - If specific visuals or cultural elements appear, choose the most context-appropriate term in \{target\_lang\} to convey the intended meaning.\\
\quad - Strive to maintain or approximate any rhymes present in the original \{source\_lang\} lyric. You may modify the literal meaning if it helps preserve rhyme and overall musicality.\\
\quad - If matching the exact syllable count is too restrictive while trying to keep it a single sentence, you are permitted to paraphrase more aggressively so that the translation remains fluid and coherent.\\
\quad - Generate the target language translation syllable list while preserving the original syllable count whenever possible. Use the audio information to synchronize this syllable list.\\
\quad - Write down the target language translation syllable list in the following format:\\
\quad\quad Example:\\
\quad\quad - Original syllable list: ["Three", "months", "of", "win", "ter", "cool", "ness"]\\
\quad - Translation syllable list: ["세", "달", "의", "겨", "울", "추", "위"] \\
\quad\quad The target syllable list has 7 syllables, maintaining the count.\\
}

{\color{orange}
3.\quad Iterate and Refine the Translation\\
\quad - After generating the initial translation, check for syllable count, natural flow, rhythm, rhyme, and meaning in context with the video and audio.\\
\quad - If the lyric was originally a single sentence, confirm that your translation remains one smooth, complete sentence in \{target\_lang\}.\\
\quad - If maintaining the syllable count as a single sentence proves too challenging, continue to refine your phrasing by paraphrasing, reordering words, or making minor adjustments to meaning. Ensure that any changes preserve naturalness and singability.\\
\quad - Repeat this process until the translation feels culturally appropriate, synchronized with the original audio, and linguistically smooth in \{target\_lang\}.\\
\quad - Before generating the final translation, please check the syllable count and the translation syllable list.\\
\quad - Write down each refined translation in the syllable list format.\\
}

{\color{olive}
4.\quad Generate the Final Translation\\
\quad - After ensuring the translation feels natural and maintains the desired structure (single sentence if the original lyric was one, etc.), use the final syllable list to form the completed translation.\\
\quad - The final translation should include natural spacing as is customary in the target language. For languages that do not typically use spacing (e.g., Chinese), do not insert additional spaces.\\
\quad - Output the final result as a single JSON in the following format:\\
\quad\quad \texttt{\{}\texttt{"translation": "final translation text"}\texttt{\}}\\
\quad\quad Example:\\
\quad\quad \texttt{\{}\texttt{"translation": "세 달의 겨울 추위"}\texttt{\}} \\
}
\quad Now, please translate the following \{source\_lang\} lyrics into \{target\_lang\} while fully complying with the above instructions.\\
\quad Real Syllable Count: \{syllable\_count\} \\
\quad Original Lyrics:
\{source\_text\}

\end{minipage}
}
\caption{\textbf{Detailed Prompt for Chain-of-Thought Lyrics Translation}}\label{tab:prompt_CoT_song_translation}
\end{table*}
\end{CJK}

\begin{CJK}{UTF8}{mj}
\begin{table*}[ht!]
\centering
\fontsize{9pt}{9.5pt}\selectfont
\noindent\fbox{
\begin{minipage}{\dimexpr\linewidth-2\fboxsep-2\fboxrule}
\tt
You are a professional song translator with expertise in preserving musicality. Translate the following song lyrics from \{source\_lang\} to \{target\_lang\}.\\
\\
- Read carefully to the provided lyrics to capture the natural rhythm, pronunciation, and any important phonetic, musical, or syllabic characteristics of the identified lyric line.\\
- You will be also given the real syllable count of the original lyric.\\
- Match the syllable count of the original lyric as closely as possible.\\
- Translate the meaning of the lyric naturally and idiomatically into the target language (\{target\_lang\}).\\
- If specific cultural elements appear, choose the most context-appropriate term in \{target\_lang\} to convey the intended meaning.\\
- If matching the exact syllable count is too restrictive while trying to keep it a single sentence, you are permitted to paraphrase more aggressively so that the translation remains fluid and coherent.\\
- Output the final result as a single JSON in the following format:\\
    ```json\\
    \{ "translation": "final translation text" \}\\
    ```\\
\\
Now, please translate the following \{source\_lang\} lyrics into \{target\_lang\} while fully complying with the above instructions.\\
\\
Real Syllable Count: \{syllable\_count\}\\
Original Lyrics:\\
\{source\_text\}
\end{minipage}
}
\caption{\textbf{Detailed Prompt for Syllable-Constraint Lyrics Translation}}\label{tab:prompt_SC_song_translation}
\end{table*}
\end{CJK}

\begin{CJK}{UTF8}{mj}
\begin{table*}[ht!]
\centering
\fontsize{9pt}{9.5pt}\selectfont
\noindent\fbox{
\begin{minipage}{\dimexpr\linewidth-2\fboxsep-2\fboxrule}
\tt
You are a professional song translator with expertise in preserving musicality. Translate the following song lyrics from {source\_lang} to {target\_lang}.\\
Please perform the following steps:\\

{\color{blue}
1. Identify the Core Lyric and Perform Syllable Segmentation\\
- Read carefully to the provided lyrics to capture the natural rhythm, pronunciation, and any important phonetic, musical, or syllabic characteristics of the identified lyric line.\\
- You will be also given the real syllable count of the original lyric. \\
- Important: You must create a syllable list that matches the syllable count of the original lyric.\\
- Break down the determined lyric line into its constituent syllables based on real syllable count.\\
- Ensure that each syllable is logically segmented according to the pronunciation.  And write it down in the following format:
- Example:\\
    - If the lyric is:\\
    `"Three months of winter coolness and awesome holidays"`\\
    - A correct segmentation might be:\\
    `["Three", "months", "of", "win", "ter", "cool", "ness"]`  \\
    - This original syllable list has 7 syllables.\\
}
\\
{\color{magenta}
2. Generate the Target Language Translation Syllable List\\
  - Translate the meaning of the lyric naturally and idiomatically into the target language (\{target\_lang\}).\\
  - If specific cultural elements appear, choose the most context-appropriate term in {target\_lang} to convey the intended meaning.\\
  - Strive to maintain or approximate any rhymes present in the original {source\_lang} lyric. You may modify the literal meaning if it helps preserve rhyme and overall musicality.\\
  - If matching the exact syllable count is too restrictive while trying to keep it a single sentence, you are permitted to paraphrase more aggressively so that the translation remains fluid and coherent.\\
  - Generate the target language translation syllable list while preserving the original syllable count whenever possible.\\
  - Write down the target language translation syllable list in the following format:\\
  - Example:\\
     - Original syllable list: `["Three", "months", "of", "win", "ter", "cool", "ness"]`  \\
     - Translation syllable list: `["세", "달", "의", "겨", "울", "추", "위"]`  \\
     - Target syllable list has 7 syllables, maintaining the count.\\
}
\\
{\color{orange}
3. Iterate and Refine the Translation\\
  - After generating the initial translation, check for syllable count, natural flow, rhythm, rhyme, and meaning in context.\\
  - If the lyric was originally a single sentence, confirm that your translation remains one smooth, complete sentence in \{target\_lang\}.\\
  - If maintaining the syllable count as a single sentence proves too challenging, continue to refine your phrasing by paraphrasing, reordering words, or making minor adjustments to meaning. Ensure that any changes preserve naturalness and singability.\\
  - Repeat this process until the translation feels culturally appropriate, and linguistically smooth in \{target\_lang\}.\\
  - Before generating the final translation, please check the syllable count and the translation syllable list.\\
  - Write down each refined translation in the syllable list format.\\
}
\\
{\color{olive}
4. Generate the Final Translation\\
  - After ensuring the translation feels natural and maintains the desired structure (single sentence if the original lyric was one, etc.), use the final syllable list to form the completed translation.\\
  - The final translation should include natural spacing as is customary in the target language. For languages that do not typically use spacing (e.g., Chinese), do not insert additional spaces.\\
  - Output the final result as a single JSON in the following format:\\
    ```json\\
    {{"translation": "final translation text"}}\\
    ``` \\
- Example:\\
    ```json\\
    {{"translation": "세 달의 겨울 추위"}}\\
    ``` \\
}
\end{minipage}
}
\caption{\textbf{Detailed Prompt for Text-Only Chain-of-Thought Lyrics Translation}}\label{tab:prompt_CoT_text_only}
\end{table*}
\end{CJK}

\section{Sample Reasoning Process for Lyrics Translation}
Table~\ref{tab:reasoning_trajectory1}, \ref{tab:reasoning_trajectory2} shows the reasoning process for Figure~\ref{fig:teaser_figure}, \ref{fig:sylavl-cot_pipeline}.
Also, Table~\ref{tab:reasoning_trajectory_al} is the reasoning process example for Appendix~\ref{appendix:cross-lingual}.
\label{appendix:reasoning_trajectory}

\begin{CJK}{UTF8}{mj}
\begin{table*}[ht!]
\centering
\fontsize{9pt}{9.5pt}\selectfont
\noindent\fbox{
\begin{minipage}{\dimexpr\linewidth-2\fboxsep-2\fboxrule}
\tt
{\color{blue}
1. Identify the Core Lyric and Perform Syllable Segmentation\\
   - The core lyric is: "And there's a butterfly".\\
   - Audio analysis indicates a natural flow with emphasis on "butterfly".\\
   - The original syllable count is 6.\\
   - Syllable segmentation: `["And", "there's", "a", "but", "ter", "fly"]`\\
}
\\
{\color{magenta}
2. Generate the Target Language Translation Syllable List Utilizing the Video Information\\
   - Video context: The scene shows a colorful, cheerful environment with a pink Troll, running happily. A butterfly flits around. The mood is light and joyful.\\
   - Translation considerations: We need a Korean translation that fits this happy, whimsical context and maintains the 6-syllable count. We can use a common Korean word for butterfly, "나비" (nabi).\\
   - Initial translation: "나비 가 있 어 요" (Na-bi ga it-eo-yo) - This translates to "There's a butterfly."\\
   - Syllable list: `["나", "비", "가", "있", "어", "요"]`\\
}
\\
{\color{orange}
3. Iterate and Refine the Translation\\
   - The initial translation "나비 가 있어요" is grammatically correct and fits the context. The syllable count is also correct at 6.\\
   - We can consider other options to see if we can get a more singable result, but this is already quite good.\\
   - Let's try "나비 가 날 아 와" (Na-bi ga nal a wa) - "A butterfly comes flying"\\
   - Syllable list: `["나", "비", "가", "날", "아", "와"]`\\
   - This also has 6 syllables and fits the scene well. It emphasizes the movement of the butterfly, which is visually present.\\
}
\\
{\color{olive}
4. Generate the Final Translation\\
   - I think "나비 가 날 아 와" is slightly better because it has a sense of movement.\\
   - Final Syllable List: `["나", "비", "가", "날", "아", "와"]`\\
   - Final Translation: 나비가 날아와\\
\\
```json\\
\{"translation": "나비가 날아와"\}\\
```\\
}
\end{minipage}
}
\caption{\textbf{Sample Reasoning Process for Chain-of-Thought Lyrics Translation}}\label{tab:reasoning_trajectory1}
\end{table*}
\end{CJK}

\begin{CJK}{UTF8}{mj}
\begin{table*}[ht!]
\centering
\fontsize{9pt}{9.5pt}\selectfont
\noindent\fbox{
\begin{minipage}{\dimexpr\linewidth-2\fboxsep-2\fboxrule}
\tt
{\color{blue}
1. Identify the Core Lyric and Perform Syllable Segmentation\\
I begin with the original lyric “Remember me, don't let it make you cry” and note the natural break after “me.” I segment the lyric into its constituent syllables as follows:\\
`["Re", "mem", "ber", "me,", "don't", "let", "it", "make", "you", "cry"]`\\
This gives a total of 10 syllables.\\
\\
}
{\color{magenta}
2. Generate the Target Language Translation Syllable List\\
The video shows a young boy singing to his elderly grandmother, who looks sad.\\ Given the context, the translation should be comforting and gentle. I will aim for a translation that conveys the same emotion and fits the rhythm of the song.\\
My initial translation is:\\
`["날", "기", "억", "해", "줘", "울", "지", "는", "마"]`\\
This translation maintains the sentiment and cultural context.\\
}
{\color{orange}
3. Iterate and Refine the Translation
Upon reviewing the segmentation ["날", "기", "억", "해", "줘", "울", "지", "는", "마"], several issues become apparent:\\
Upon reviewing the initial segmentation ["날", "기", "억", "해", "줘", "울", "지", "는", "마"], the primary issue is the syllable count.\\
The list contains 9 syllables, not 10 as originally noted. This mismatch is crucial for accurately fitting the lyric to the song's original melody and rhythm, which is built on 10 syllables.\\
Furthermore, the original lyric has a distinct 4+6 syllable structure with a natural break after the 4th syllable ("me,"). The initial translation has a 5+4 structure (["날", "기", "억", "해", "줘"] as the first part and ["울", "지", "는", "마"] as the second), causing the potential break point to fall after the 5th syllable instead of the 4th.\\
This misalignment in both total syllable count and internal structure will disrupt the timing and flow of the lyric when sung to the original music, making it difficult to match the notes and rhythm.\\
\\
Second attempt (adjusting for syllable count and naturalness):\\
`["날", "잊", "지", "마", "슬", "퍼", "하", "지", "는", "마"]`\\
The second attempt seems more natural and maintains the 10-syllable count.\\ Let's check the meaning and flow. It translates roughly to "Don't forget me, don't be sad". This captures the essence of the original lyric.\\
}
{\color{olive}
4. Generate the Final Translation
The final syllable list is:\\
`["날", "잊", "지", "마", "슬", "퍼", "하", "지", "는", "마"]`\\
\\
```json\\
\{"translation": "날 잊지 마 슬퍼하지는 마"\}\\
```\\
}
\end{minipage}
}
\caption{\textbf{Sample Reasoning Process for Chain-of-Thought Lyrics Translation}}\label{tab:reasoning_trajectory2}
\end{table*}
\end{CJK}

\begin{CJK}{UTF8}{mj}
\begin{table*}[ht!]
\centering
\fontsize{9pt}{9.5pt}\selectfont
\noindent\fbox{
\begin{minipage}{\dimexpr\linewidth-2\fboxsep-2\fboxrule}
\tt
{\color{blue}
1. Identify the Core Lyric and Perform Syllable Segmentation\\
The original lyric is “기억해 줘 지금 떠나가지만”.\\
I segment the lyric into its constituent syllables as follows:\\
`["기", "억", "해", "줘", "지", "금", "떠", "나", "가", "지", "만"]`\\
This gives a total of 11 syllables.\\
}
\\
{\color{magenta}
2. Generate the Target Language Translation Syllable List\\
The video shows a young boy singing to an elderly woman in a wheelchair. The boy is trying to cheer her up with the song.\\
Given the context, I need to translate the line to convey a sense of longing and remembrance as the singer departs. I aim to match the syllable count and create a natural flow.\\
My initial translation is:\\
`["Re", "mem", "ber", "me", "though", "I", "have", "to", "leave", "you", "now"]`\\
}
\\
{\color{orange}
3. Iterate and Refine the Translation\\
The initial translation has 11 syllables, which matches the original. The phrase "remember me" captures the essence of the original lyric "기억해 줘". The phrase "though I have to leave you now" translates "지금 떠나가지만" effectively, while maintaining a natural rhythm.\\
}
\\
{\color{olive}
4. Generate the Final Translation\\
The translation meets all the required conditions: natural flow, correct syllable count, and cultural appropriateness.\\
The final English translation is:\\
"Remember me though I have to leave you now"\\
\\
```json\\
\{\\
"translation": "Remember me though I have to leave you now"\\
\}\\
```\\
}
\end{minipage}
}
\caption{\textbf{Sample Reasoning Process for Chain-of-Thought Lyrics Translation From Korean to English}}\label{tab:reasoning_trajectory_al}
\end{table*}
\end{CJK}

\section{Implementation Details}
To evaluate and analyze multilingual lyrics, we conducted preprocessing for multilingual data. Since preprocessing must be tailored to the linguistic characteristics of each language, we utilized various libraries, as shown in Table~\ref{tab:implement_lib}. After this process, to account for syllables in numbers, we used num2words~\footnote{https://github.com/savoirfairelinux/num2words} to convert numerical values into words, followed by IPA transcription~\cite{Mortensen-et-al:2018} and syllable counting.

\begin{table}[htbp]
\resizebox{\linewidth}{!}{%
\begin{tabular}{ccc}
\toprule
& Syllable& IPA\\ \midrule
EN & Syllabifier~\footnote{https://github.com/anson-vandoren/syllabifier} & epitran~\footnote{https://github.com/dmort27/epitran} \\
ES & count-syllables-in-spanish~\footnote{https://github.com/pablolucianop/count-syllables-in-spanish} & epitran \\
FR & syllabify-fr~\footnote{https://github.com/UrielCh/syllabify-fr} & epitran \\
KO & length of text& epitran \\
JA & japanese-mora-counter~\footnote{https://github.com/grocio/japanese-mora-counter} & epitran \\
\bottomrule
\end{tabular}%
}
\caption{\textbf{Libraries used for syllable counting and IPA conversion across languages.}}\label{tab:implement_lib}
\end{table}

\section{Experiment Details}
\textbf{Computational Resources} \\
Translating lyrics with Qwen-72B took up to 24 hours, when using 16 NVIDIA A6000s.\\
the average number of API calls to translate all lines in the dataset is approximately 20,000, which costs about \$400K total for Gemini 2.0 Flash.\\
For all language cross-lingual translation, the number of API calls were about 40,000.

For the generation settings used in Gemini and Qwen, please refer to Table~\ref{tab:gemini_config} and Table~\ref{tab:qwen_config}, respectively.
\begin{table}[ht]
\centering
\begin{tabular}{ll}
\toprule 
\textbf{Parameter} & \textbf{Value} \\
\midrule 
\texttt{temperature} & 0.6 \\
\texttt{top\_p} & 0.95 \\
\texttt{top\_k} & 40 \\
\texttt{max\_output\_tokens} & 8192 \\
\texttt{response\_mime\_type} & \texttt{text/plain} \\
\bottomrule 
\end{tabular}
\caption{\textbf{Gemini generation configuration.}}
\label{tab:gemini_config}
\end{table}

\begin{table}[ht]
\centering
\begin{tabular}{ll}
\toprule 
\textbf{Parameter} & \textbf{Value} \\
\midrule 
\texttt{temperature} & 0.7 \\
\texttt{top\_p} & 0.8 \\
\texttt{max\_tokens} & 4096 \\
\texttt{presence\_penalty} & 1.05 \\
\bottomrule 
\end{tabular}
\caption{\textbf{Qwen generation configuration.}}
\label{tab:qwen_config}
\end{table}

\end{document}